\documentclass[letterpaper, 10 pt, conference]{ieeeconf}  

\IEEEoverridecommandlockouts                              

\usepackage{makecell} 
\usepackage{caption}
\usepackage{subcaption}
\usepackage{amsmath,amssymb} 
\usepackage{color}
\usepackage{graphicx}
\usepackage{flushend}

\definecolor{multi_trans}{RGB}{192,31,219}
\definecolor{pedformer}{RGB}{243,144,11}
\definecolor{sep_dec}{RGB}{19,76,227}
\definecolor{no_int}{RGB}{22,222,232}
\definecolor{gt}{RGB}{16,244,15}

\usepackage{pifont}
 \usepackage[normalem]{ulem}
 \useunder{\uline}{\ul}{}

\usepackage{float}
\usepackage{hyperref}       
\usepackage{multirow}
\graphicspath{{images/}}
\usepackage{array}
\newcolumntype{?}{!{\vrule width 1pt}}
\title{\LARGE \bf
A Novel Benchmarking Paradigm and a Scale- and Motion-Aware Model for Egocentric Pedestrian Trajectory Prediction
}

\author{Amir Rasouli 
\thanks{Huawei Technologies Canada   {\tt\small amir.rasouli@huawei.com}.
        }}

\begin{document}

\maketitle
\thispagestyle{empty}
\pagestyle{empty}

\begin{abstract}
Predicting pedestrian behavior is one of the main challenges for intelligent driving systems. In this paper, we present a new paradigm for evaluating egocentric pedestrian trajectory prediction algorithms. Based on various contextual information, we extract driving scenarios for a meaningful and systematic approach to identifying challenges for prediction models. In this regard, we also propose a new metric for more effective ranking within the scenario-based evaluation. We conduct extensive empirical studies of existing models on these scenarios to expose shortcomings and strengths of different approaches. The scenario-based analysis highlights the importance of using multimodal sources of information and challenges caused  by inadequate modeling of ego-motion and scale of pedestrians. To this end, we propose a novel egocentric trajectory prediction model that benefits from multimodal sources of data fused in an effective and efficient step-wise hierarchical fashion and two auxiliary tasks designed to learn more robust representation of scene dynamics. We show that our approach achieves significant improvement by up to 40\% in challenging scenarios compared to the past arts via empirical evaluation on common benchmark datasets.
\end{abstract}

\section{Introduction}
Pedestrian behavior prediction requires modeling of various contextual information, such as scene dynamics, pedestrian state, etc. \cite{Rasouli_2017_IV,Rasouli_2019_ITS}. Prediction models are often evaluated on benchmarks by computing their performance on the entire driving datasets. However, due to inherent biases in such datasets, e.g., prevalence of cruising in AD datasets \cite{Chang_2019_CVPR} or signalized intersections in pedestrian datasets \cite{Rasouli_2019_ICCV}, and high diversity of real-world driving scenarios, such a high-level benchmarking says little about the robustness of the algorithms to different challenges arising from traffic scenes.

\begin{figure}[!t]
\vspace{0.2cm}
\centering
\includegraphics[width=1\columnwidth]{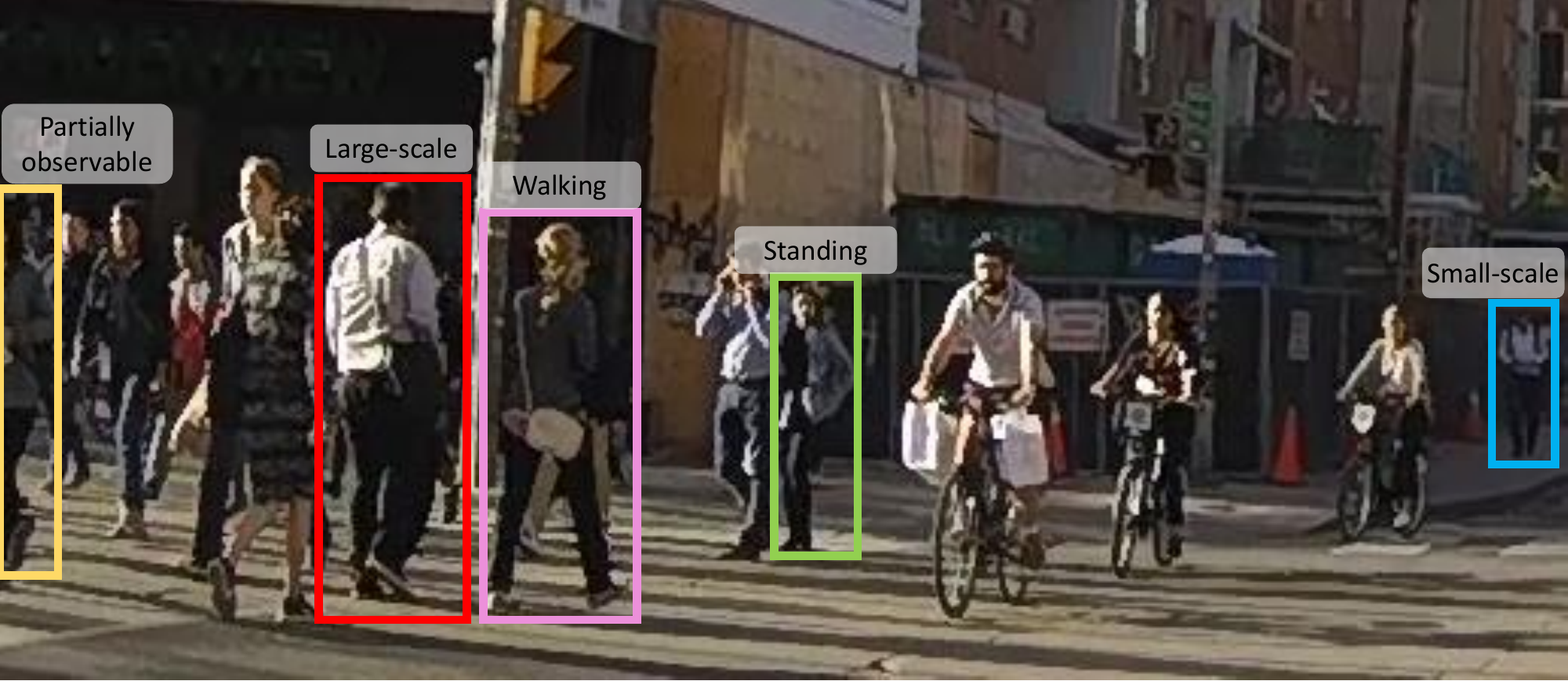}
\caption{Factors, such as observability, state, and scale, impact pedestrian trajectory prediction in an egocentric setting.}
\label{fig:first_image}
\vspace{-0.6cm}
\end{figure}

To this end,  we propose a new paradigm for evaluating egocentric pedestrian trajectory prediction models. Our goal is to identify factors suitable for extracting scenarios that expose various challenges for the prediction algorithms (see Figure \ref{fig:first_image}). We propose an effective metric to rank the performance of the models in scenario-based analysis, and present an extensive evaluation of the existing models using the proposed evaluation scheme. Based on our findings from the evaluation, we propose a novel prediction algorithm that achieves state-of-the-art performance on common benchmark datasets. We show the effectiveness of our model by empirical study on common benchmarks and ablation study.
\vspace{-0.15cm}

\section{Related Works} 
\subsection{Egocentric Pedestrian Trajectory Prediction}
Pedestrian trajectory prediction is usually done either from a bird's eye view \cite{yuan2021agentformer, Dendorfer_2021_ICCV, Shafiee_2021_CVPR, Hu_2020_CVPR, Mohamed_2020_CVPR,Sun_2020_CVPR,Sun_2020_CVPR_2,Mangalam_2020_ECCV, Choi_2019_ICCV,Zhang_2019_CVPR,Sadeghian_2019_CVPR,Gupta_2018_CVPR} or egocentric \cite{damirchi2023context,li2023icra, halawa2022action, wang2022stepwise, yao2021bitrap, Neumann_2021_CVPR, Makansi_2020_CVPR, Malla_2020_CVPR,Rasouli_2019_ICCV,Yagi_2018_CVPR, Yao_2019_ICRA,Yao_2019_IROS,Bhattacharyya_2018_CVPR, Chandra_2019_CVPR} perspective. The former is applied to scenes recorded from a fixed surveillance camera perspective or the projection of driving scenes into a global coordinate system. Egocentric prediction involves recordings from a moving egocentric camera view, hence the scale of the objects in the scenes may change and the observed motion is a combination of ego-motion and motion of other dynamic objects. To model these factors, many algorithms rely on the apparent changes in the scale of the pedestrians, e.g., the size of bounding boxes around them, and implicitly reason about the scene dynamics \cite{wang2022stepwise,yao2021bitrap}. Other models use optical flow information \cite{Yao_2019_ICRA} or train models based on pedestrian states, e.g., walking or standing \cite{halawa2022action}. However, without explicit ego-motion information, these models are limited in identifying whether the observed motion is caused by the ego-vehicle, pedestrian, or both.

Some models use more explicit information, such as ego-speed \cite{Rasouli_2021_ICCV, Rasouli_2019_ICCV} or angular velocity \cite{rasouli2023pedformer} along with contextual information, such as scene semantics and interactions \cite{rasouli2023pedformer, Rasouli_2021_ICCV} or intentions \cite{Rasouli_2019_ICCV}, often combined in a multimodal framework that also comes with the added challenge of fusing information from different sources. In the trajectory prediction domain, common approaches include early \cite{damirchi2023context} or late fusion \cite{Rasouli_2019_ICCV}, bimodal fusion \cite{Rasouli_2021_ICCV}, hierarchical fusion \cite{nayakanti2023wayformer}, and pair-wise cross-modal fusion \cite{rasouli2023pedformer}. However, they either lack the ability to effectively capture the correlation between different data modalities or are computationally expensive. We propose an efficient model that effectively fuses different sources of information in a \textit{step-wise} hierarchical fashion that effectively captures the correlation between different data modalities while minimizing the computational overhead.

\subsection{Benchmarking and Evaluation} 
There are many egocentric driving datasets, some of which are specifically catered to pedestrian prediction \cite{Rasouli_2017_ICCVW, Rasouli_2019_ICCV, Malla_2020_CVPR,  Girase_2021_ICCV, Yao_2019_ICRA, Liu_2020_RAL}. These datasets are often collected in urban environments and provide annotations for various road structures and contextual factors, e.g., signals, pedestrian behavior, etc. Given that these datasets are collected in natural settings, the recordings are often biased towards simpler driving scenarios, such as driving straight, interacting with pedestrians at signalized intersections, driving at constant speed, etc. Benchmarking on entire datasets with such characteristics, at best, provides a general ranking but reveals very little about common challenges for prediction or models' characteristics. Hence, a more detailed scenario-based evaluation paradigm is needed.

\noindent \textbf{Contributions} of our paper are,
1) We propose a new scenario-based paradigm for evaluating egocentric pedestrian trajectory prediction models and a new metric for better ranking of the models under each scenario. 2) We conduct extensive evaluation of existing methods on the  scenarios and provide insights into common challenges for the algorithms. 3) Based on our findings, we propose a novel state-of-the-art model that takes advantage of multimodal data input and auxiliary tasks to learn more robust representation of the scene dynamics and small-scale samples, two of common challenges for egocentric prediction models. 4) Lastly, we present experimental results on the proposed model on common benchmark datasets followed by ablation studies.

\section{Scenario-based Benchmark}
\vspace{-0.1cm}
\subsection{Problem Formulation}
We formulate egocentric trajectory prediction as an optimization problem, the goal of which is to learn a distribution $p(L|C)$ where $L = \{l^{t+1:t+\tau}_i\}$ is the trajectory of a pedestrian $i$,  $1 < i < n$, and $ C = \{c^{t-o+1:t}_i\}$ denotes observation context. Here,  $o$ and $\tau$ correspond to observation and prediction horizons, respectively. In this setting, each point is in the form of $[x_1,y_1,x_2,y_2]$ which captures the coordinates of bounding box around the pedestrian.

\subsection{Scenario Extraction} \label{sec:scenarios}
The first step for effective benchmarking is to divide the evaluation dataset into meaningful subsets that highlight different aspects of the models. Unlike bird's eye view prediction, in the egocentric setting, coordinates are in image plane, hence they may vary depending on motion of both the ego-camera and pedestrians.

\subsubsection{Factors} To characterize the subsets of the data, we look at both pedestrians and ego-camera factors. \textbf{Pedestrian} factors are as follows: \textit{scale}, which reflects the proximity of pedestrians to the ego-camera. We use the height of bounding boxes as a measure of scale, since width of boxes can fluctuate due to pedestrian gait. \textbf{State}, which refers to whether the pedestrian is walking or standing. Predicting the future trajectory of a walking pedestrian in the presence of ego-motion can be challenging because there are two sources of motion combined. \textbf{Ego-camera} is described in terms of ego-\textit{speed} (km/h) and ego-\textit{action} (going straight/turning).

\subsubsection{Changing behavior} One of the key challenges in the context of prediction is when the behavior of agents is different within the observation and prediction horizons. For instance, pedestrians might be standing during observation period and starts walking during prediction period and vice versa. Similarly, the speed of the ego-camera can vary across observation and prediction horizons. Such changing behavior can potentially pose a challenge for prediction models. 

\subsection{Trajectory Metrics}
Distance-based metrics are most common for egocentric prediction \cite{halawa2022action, wang2022stepwise, yao2021bitrap, Neumann_2021_CVPR}, where the error is calculated as the mean square error (MSE) of bounding boxes or their center coordinates. Although effective, these metrics are not adequate for scenario-based analysis since bounding box scales change depending on pedestrians' proximity to the ego-camera (see Figure \ref{fig:first_image}), causing the larger scale boxes' error to dominate the overall metric value. To reflect the real-world changes, error term should be measured relative to the scale of the pedestrian. This is due to the perspective effect in the image plane, causing the same pixel difference values to correspond to very different real-world values depending on the proximity of the pedestrian to the ego-camera.

\textbf{Scaled Distance Error} is a new metric proposed for our scenario-based evaluation. We compute the metric by scaling the pixel error values by the average area of bounding boxes, computed based on their widths and heights.  

\textit{Boundary issue}. The full area of bounding boxes is not always visible due to occlusion or as a result of  pedestrians entering/exiting the scene, as shown in the leftmost sample in Figure \ref{fig:first_image}. Hence, simply computing areas based on the visible boxes results in incorrect scaling of certain samples. To address this issue, we first measure the average aspect ratio of height and width of the fully visible bounding boxes in the dataset. We then use this ratio to identify the samples that are partially observable and then adjust their ratio accordingly.

\section{The Benchmark}
\begin{table*}[t!]
\vspace{0.2cm}
\caption{Benchmark results on single-factor scenarios. The third row indicates the number of train and test samples scaled by $10^3$ and results are reported as $B_{mse}$/$sB_{mse}$.}
\label{tbl:single_factor_ped}
\resizebox{\textwidth}{!}{%
\begin{tabular}{llllllllll}
 &
  \multicolumn{7}{c|}{\textbf{Pedestrian Scale (pixels)}} &
  \multicolumn{2}{c}{\textbf{Pedestrian State}} \\  \cline{2-10} 
 &
  \multicolumn{1}{c|}{\textit{0-50}} &
  \multicolumn{1}{c|}{\textit{50-80}} &
  \multicolumn{1}{c|}{\textit{80-100}} &
  \multicolumn{1}{c|}{\textit{100-150}} &
  \multicolumn{1}{c|}{\textit{150-200}} &
  \multicolumn{1}{c|}{\textit{200-300}} &
  \multicolumn{1}{c|}{\textit{300+}} &
  \multicolumn{1}{c|}{\textit{Walking}} &
  \multicolumn{1}{c}{\textit{Standing}} \\ \cline{2-10}
  & 
  \multicolumn{1}{c|}{\textit{5.2/5.3}} &
  \multicolumn{1}{c|}{\textit{10.8/8.6}} &
  \multicolumn{1}{c|}{\textit{4.9/4.4}} &
  \multicolumn{1}{c|}{\textit{8.2/5.7}} &
  \multicolumn{1}{c|}{\textit{5.4/4.8}} &
  \multicolumn{1}{c|}{\textit{6.5/4.6}} &
  \multicolumn{1}{c|}{\textit{3.4/2.9}} &
  \multicolumn{1}{c|}{\textit{22.4/18.3}} &
  \multicolumn{1}{c}{\textit{20.0/16.1}} \\ \hline
  \multicolumn{1}{l|}{PIE$_{traj}$} &
  \multicolumn{1}{c|}{187/0.308} &
  \multicolumn{1}{c|}{170/0.105} &
  \multicolumn{1}{c|}{373/0.122} &
  \multicolumn{1}{c|}{595/0.098} &
  \multicolumn{1}{c|}{709/0.058} &
  \multicolumn{1}{c|}{980/0.041} &
  \multicolumn{1}{c|}{2064/0.036} &
  \multicolumn{1}{c|}{677/0.048} &
  441/0.060 \\
\multicolumn{1}{l|}{B-LSTM} &
  \multicolumn{1}{c|}{225/0.371} &
  \multicolumn{1}{c|}{215/0.132} &
  \multicolumn{1}{c|}{464/0.152} &
  \multicolumn{1}{c|}{840/0.138} &
  \multicolumn{1}{c|}{886/0.073} &
  \multicolumn{1}{c|}{1170/0.049} &
  \multicolumn{1}{c|}{2492/0.044} &
  \multicolumn{1}{c|}{833/0.059} &
  582/0.080  \\
\multicolumn{1}{l|}{PIE$_{full}$} &
  \multicolumn{1}{c|}{195/0.322} &
  \multicolumn{1}{c|}{163/0.101} &
  \multicolumn{1}{c|}{{\ul 313/0.102}} &
  \multicolumn{1}{c|}{{\ul 447/0.074}} &
  \multicolumn{1}{c|}{{\ul 513/0.042}} &
  \multicolumn{1}{c|}{{\ul 628/0.026}} &
  \multicolumn{1}{c|}{1838/0.032} &
  \multicolumn{1}{c|}{{\ul 559/0.040}} &
  {\ul 337/0.046}\\
\multicolumn{1}{l|}{BiTraP} &
  \multicolumn{1}{c|}{{\ul 134/0.220}} &
  \multicolumn{1}{c|}{{\ul 158/0.097}} &
  \multicolumn{1}{c|}{361/0.118} &
  \multicolumn{1}{c|}{538/0.088} &
  \multicolumn{1}{c|}{661/0.054} &
  \multicolumn{1}{c|}{852/0.036} &
  \multicolumn{1}{c|}{{\ul 1633/0.029}} &
  \multicolumn{1}{c|}{579/0.041} &
  400/0.055\\
\multicolumn{1}{l|}{PedFormer} &
  \multicolumn{1}{c|}{\textbf{76/0.125}} &
  \multicolumn{1}{c|}{\textbf{72/0.044}} &
  \multicolumn{1}{c|}{\textbf{124/0.041}} &
  \multicolumn{1}{c|}{\textbf{250/0.041}} &
  \multicolumn{1}{c|}{\textbf{295/0.024}} &
  \multicolumn{1}{c|}{\textbf{522/0.022}} &
  \multicolumn{1}{c|}{\textbf{1363/0.024}} &
  \multicolumn{1}{c|}{\textbf{394/0.028}} &
  \textbf{153/0.021}
 \end{tabular}
 \vspace{-0.8cm}
}
\vspace{-0.5cm}
\end{table*} 

\begin{table}[t!]
\caption{Benchmark results on single-factor scenarios. The third row indicates the number of train and test samples scaled by $10^3$ and results are reported as $B_{mse}$/$sB_{mse}$.}
\label{tbl:single_factor_veh}
\resizebox{\columnwidth}{!}{%
\begin{tabular}{lllllll}
 &
  \multicolumn{6}{c}{\textbf{Ego-speed (km/h)}} \\ \cline{2-7} 
 &
  \multicolumn{1}{c|}{\textit{0}} &
  \multicolumn{1}{c|}{\textit{0-5}} &
  \multicolumn{1}{c|}{\textit{5-10}} &
  \multicolumn{1}{c|}{\textit{10-20}} &
  \multicolumn{1}{c|}{\textit{20-30}} &
  \multicolumn{1}{c}{\textit{30+}} \\ \cline{2-7} 
 &
  \multicolumn{1}{c|}{\textit{20.2/19.7}} &
  \multicolumn{1}{c|}{\textit{6.8/4.8}} &
  \multicolumn{1}{c|}{\textit{5.1/2.4}} &
  \multicolumn{1}{c|}{\textit{6.7/4.0}} &
  \multicolumn{1}{c|}{\textit{3.6/3.5}} &
  \multicolumn{1}{c}{\textit{2.0/1.8}} \\ \hline
  \multicolumn{1}{l|}{PIE$_{traj}$} &
  \multicolumn{1}{c|}{284/0.020} &
  \multicolumn{1}{c|}{800/0.071} &
  \multicolumn{1}{c|}{1495/0.136} &
  \multicolumn{1}{c|}{1153/0.189} &
  \multicolumn{1}{c|}{709/0.208} &
  649/0.320 \\
\multicolumn{1}{l|}{B-LSTM} &
  \multicolumn{1}{c|}{367/0.026} &
  \multicolumn{1}{c|}{884/0.079} &
  \multicolumn{1}{c|}{1849/0.168} &
  \multicolumn{1}{c|}{1405/0.231} &
  \multicolumn{1}{c|}{954/0.280} &
  969/0.477 \\
\multicolumn{1}{l|}{PIE$_{full}$} &
  \multicolumn{1}{c|}{{\ul 259/0.019}} &
  \multicolumn{1}{c|}{{\ul 659/0.059}} &
  \multicolumn{1}{c|}{{\ul 939/0.085}} &
  \multicolumn{1}{c|}{{\ul 976/0.160}} &
  \multicolumn{1}{c|}{{\ul 563/0.165}} &
  {\ul 317/0.156} \\
\multicolumn{1}{l|}{BiTraP} &
  \multicolumn{1}{c|}{{\ul 269/0.019}} &
  \multicolumn{1}{c|}{704/0.063} &
  \multicolumn{1}{c|}{1221/0.111} &
  \multicolumn{1}{c|}{1018/0.167} &
  \multicolumn{1}{c|}{597/0.175} &
  386/0.190 \\
\multicolumn{1}{l|}{PedFormer} &
  \multicolumn{1}{c|}{\textbf{225/0.016}} &
  \multicolumn{1}{c|}{\textbf{390/0.035}} &
  \multicolumn{1}{c|}{\textbf{487/0.044}} &
  \multicolumn{1}{c|}{\textbf{467/0.077}} &
  \multicolumn{1}{c|}{\textbf{272/0.080}} &
  \textbf{225/0.111}
\end{tabular}
 \vspace{-1cm}
}
\vspace{-0.5cm}
\end{table}

\textbf{Dataset.} We use the publicly available \textit{Pedestrian Intention Estimation (PIE)} \cite{Rasouli_2019_ICCV} dataset that contains a diverse set of labels for pedestrians and ego-vehicle. The data consists of $6$ hours of recording from a monocular camera inside a moving vehicle. Following the common protocol for egocentric prediction \cite{wang2022stepwise, yao2021bitrap, Rasouli_2019_ICCV}, we extract sequences with $50\%$ overlap divided into $0.5s$ for observation and $1.5s$ prediction. The evaluations are done using the default test set.

\textbf{Scenarios.} We extract scenarios based on the factors described in Sec. \ref{sec:scenarios}. Using the height of pedestrians,  we divide the data by scale into categories with balanced number of samples. For the state, we use walking and standing labels and consider the majority voting to characterize the observation/prediction sequences as either walking or standing. For ego, we consider the vehicle's speed to characterize ego-motion, yaw angle with the threshold of $5^o$ change to determine turning action, and acceleration threshold of $0.3m/s^2$ to determine speed changes.

\textbf{Metrics.} We report on the commonly used metrics for egocentric trajectory prediction \cite{wang2022stepwise,Rasouli_2019_ICCV}. For the sake of brevity, we only report on MSE error over bounding boxes ($B_{MSE}$) and its proposed scaled version denoted as $sB_{MSE}$ averaged over the entire prediction horizon. The remaining metrics will be released online upon the acceptance of the paper. 

\textbf{Models.} Our goal is to highlight the differences between models that rely on different architectures, learning methods and contextual information. For this purpose, we report on the following models: \textbf{PIE$_{traj}$} \cite{Rasouli_2019_ICCV} which is a basic recurrent encoder-decoder architecture that only uses bounding box coordinates; \textbf{PIE$_{full}$} which is an extension of PIE$_{traj}$ that also incorporates pedestrian intention and ego-speed; \textbf{B-LSTM} \cite{Bhattacharyya_2018_CVPR}, a recurrent architecture with Bayesian weight learning method; \textbf{BiTraP} \cite{yao2021bitrap}, a conditional variational (CVAE) model that uses pedestrian goals and a bidirectional decoder for prediction; and \textbf{PedFormer} \cite{rasouli2023pedformer}, a multitasking framework with hybrid Transformer-recurrent architecture\footnote{Since trajectory samples are extracted over the entire tracks (not up to crossing events), we drop the action prediction branch from PedFormer and only predict trajectories and grid locations.} 

\subsection{Single-factor Scenarios}\label{sec:sing-factor}

Here, we extract scenarios only based on a single factor, namely the scale of pedestrians, their state, and ego-speed. For pedestrian states, we only report on the cases where the state of the pedestrians is the same across observation and prediction. The results are shown in Table \ref{tbl:single_factor_ped}.

When considering the absolute error, categories corresponding to larger scales appear worse. However, once the error is scaled, we can see that relatively speaking, prediction of smaller scale pedestrians are more challenging. Although performance degradation happens for all models, the rate of change varies and results in different ranking across different scenarios for PIE$_{full}$ and BiTraP.

As expected, pedestrian state also impacts the performance. Intuitively, one would expect scaled error to be lower when the pedestrian is standing. However, this is not the case for most models due to other confounding factors that are not considered under state scenarios. This means that further breakdown of scenarios with additional factors is needed to get a better understanding of the performance variation. 

A similar trend of changes in performance can be seen in different speed categories, as shown in Table \ref{tbl:single_factor_veh}. Here, we can see that at speed 0, where no motion is present, the top three models perform within a similar range. As the ego-speed increases, the gap between the models with ego-motion modeling and others increases significantly. It should be noted that the way ego-motion is modeled is also important. For instance, PIE$_{full}$ only uses the absolute speed of the ego-vehicle which does not provide much added benefit compared to BiTraP which does not use the speed. For PedFormer, on the other hand, the performance gap is more as it also uses the angular velocity of ego-vehicle.

\subsection{Two-factor Scenarios}
\label{sec:two_factor}
\begin{figure}[!t]
\vspace{0.2cm}
\centering
\includegraphics[width=1\columnwidth]{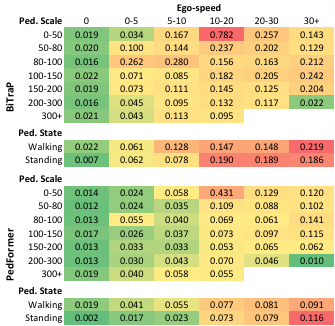}
\caption{Two-factor scenario-based evaluation using $sB_{MSE}$.}
\label{fig:2_factor}
\vspace{-0.5cm}
\end{figure}

We saw that single-factor analysis is not always sufficient. Hence, we extract scenarios based on combinations of two factors, namely scale and ego-speed, as well as state and ego-speed. Following the previous experiment, since our focus is mainly on scale changes and due to space limitations, we report the results only on $sB_{MSE}$ metric.

The results are illustrated in Figure \ref{fig:2_factor}. As expected, the performance of both models degrades as the speed increases (row-wise) and improves as scale increases (column-wise). The rate of degradation, however, is different. For BiTraP, which does not explicitly model ego-motion, the performance significantly drops in the presence of even lower speed. 

Similar degradation is observable in the state vs. speed case. An interesting finding is that the ranking between walking and standing, within each speed category, differs across models. For PedFormer, walking  samples are more challenging, as one would expect, all the way up to 30+ speed category where the ranking flips. For BiTraP, on the other hand, the ranking switches once at 10-20 category and again at 30+. One potential reason can be the relative motion between walking pedestrians and ego-vehicle. As the speed gets higher, pedestrian motion may appear smaller or even constant in image plane, and as a consequence in the case of PedFormer, performance degradation is reduced as the model relies more on the provided ego-motion information. This also applies to BiTraP but only up to a certain degree, hence the ranking fluctuates more. 

In two-factor evaluation, we can see some anomalies that do not follow the overall performance trends. For instance, for both BiTraP and PedFormer, there is a significant performance drop at scale 0-50 and speed 10-20. Such drastic changes can be due to additional factors in those categories, e.g., pedestrian state, ego-vehicle action, abrupt changes between observation and prediction, etc. To shed more light on it, one can use additional factors to extract finer scenario categories. However, it may lead to a longtail effect, i.e., small number of training samples in each scenario category. For instance, in the proposed two-factor scenarios, there are no instances of high-speed driving and large scales, and in some categories the number of samples is below 100. It should also be noted that some anomalies are model-specific, e.g., BiTraP unexpectedly does very well at scale 0-50 and speed 0-5.

\subsection{Challenging Scenarios}

\begin{table}[t!]
\vspace{0.2cm}
\caption{The performance of models on challenging scenarios. $W$ and $S$ refer to walking and standing respectively and the transition of one state to other from observation $_o$ to prediction $_p$.  The third row indicates the number of train and test samples by $10^3$ and results are reported as $B_{mse}$/$sB_{mse}$.}
\label{challenging_scnearios}
\resizebox{\columnwidth}{!}{%
\begin{tabular}{lcc|cc|cc}
 &
  \multicolumn{2}{c|}{\textbf{Pedestrian State}} &
  \multicolumn{2}{c|}{\textbf{Ego-motion}} &
  \multicolumn{2}{c}{\textbf{Ego-action}} \\ \cline{2-7} 
 &
  \multicolumn{1}{c|}{\textbf{$W_o$-$S_p$}} &
  \multicolumn{1}{c|}{\textbf{$S_o$-$W_p$}} &
  \multicolumn{1}{c|}{\textbf{Constant}} &
  \multicolumn{1}{c|}{\textbf{Change}} &
  \multicolumn{1}{c|}{\textbf{Straight}} &
  \multicolumn{1}{c}{\textbf{Turn}} \\ \cline{2-7} 
 &
  \multicolumn{1}{c|}{\textit{1.2/1.0}} &
  \multicolumn{1}{c|}{\textit{1.0/0.8}} &
  \multicolumn{1}{c|}{\textit{41.5/34.2}} &
  \multicolumn{1}{c|}{\textit{3.0/2.0}} &
  \multicolumn{1}{c|}{\textit{41.3/34.7}} &
  \multicolumn{1}{c}{\textit{3.2/1.5}} \\ \hline
\multicolumn{1}{l|}{PIE$_{traj}$} &
  \multicolumn{1}{c|}{536/0.064} &
  1553/0.108 &
  \multicolumn{1}{c|}{474/0.043} &
  2576/0.236 &
  \multicolumn{1}{c|}{436/0.040} &
  4166/0.430 \\
\multicolumn{1}{l|}{B-LSTM} &
  \multicolumn{1}{c|}{513/0.061} &
  1847/0.128 &
  \multicolumn{1}{c|}{593/0.054} &
  3213/0.295 &
  \multicolumn{1}{c|}{542/0.049} &
  5274/0.545 \\
\multicolumn{1}{l|}{PIE$_{full}$} &
  \multicolumn{1}{c|}{{\ul 376/0.045}} &
  1167/0.081 &
  \multicolumn{1}{c|}{{\ul 402/0.037}} &
  {\ul 1636/0.150} &
  \multicolumn{1}{c|}{{\ul 338/0.031}} &
  3547/0.366 \\
\multicolumn{1}{l|}{BiTraP} &
  \multicolumn{1}{c|}{499/0.059} &
  {\ul 1142/0.079} &
  \multicolumn{1}{c|}{430/0.039} &
  1899/0.174 &
  \multicolumn{1}{c|}{383/0.035} &
  {\ul 3492/0.361} \\
\multicolumn{1}{l|}{PedFormer} &
  \multicolumn{1}{c|}{\textbf{252/0.030}} &
  \textbf{970/0.067} &
  \multicolumn{1}{c|}{\textbf{259/0.024}} &
  \textbf{924/0.085} &
  \multicolumn{1}{c|}{\textbf{248/0.023}} &
  \textbf{1422/0.147}
\end{tabular}
\vspace{-0.5cm}
}\vspace{-0.5cm}
\end{table}

Table \ref{challenging_scnearios} summarizes the results of experiments on challenging scenarios. As discussed earlier, changes in the state of pedestrians and the ego-vehicle can have a significant negative impact on prediction. One interesting observation here is the significant difference between $W_o-S_p$ and $S_o-W_p$ cases. The reason for this can be that generally in $W_o-S_p$ cases, the pedestrian is slowing down, indicating the possibility of standing in the near future. In $S_o-W_p$ cases, however, there is no motion history to hint the possibility of walking in the future. Thus, the models require a better understanding of the context (other than dynamics) to infer pedestrians' future state. 

Lastly, we can see a drastic difference between ego-vehicle moving straight vs. turning. Turn actions often generate irregular motion patterns, which are difficult to estimate without a clear understanding of the road structure or the state of other agents. Including information, such as angular velocity, can help improve the results (as in PedFormer) but are still not sufficient for an accurate estimation. 

\section{Model}
\begin{figure}[!t]
\vspace{0.2cm}
\centering
\includegraphics[width=1\columnwidth]{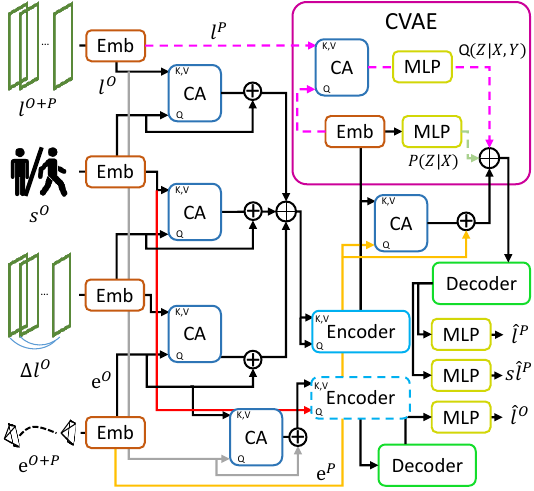}
\caption{Model architecture where pink dashed lines indicate the flow during training and green dashed line indicates test time. O and P stand for observation and prediction horizons.}
\label{fig:diagram}\vspace{-0.5cm}
\end{figure}
Based on the findings of the scenario-based analysis, in this section, we propose a novel model for egocentric prediction. In this model, we emphasize modeling dynamic factors to improve the overall performance, as illustrated in Figure \ref{fig:diagram}. The proposed model is a fully attention-based architecture divided into three main modules: a multimodal encoder, decoders, and an input reconstructor.

\subsection{Multimodal Encoder}
For capturing scene dynamics, we rely on a set of multimodal inputs, namely pedestrian normalized locations, states (actions) as an indicator of whether pedestrian is moving, velocity computed based on the changes in location of pedestrian at time $t$, and the ego-motion, which includes GPS coordinates, acceleration, and angular velocity.

As demonstrated by past arts \cite{rasouli2023pedformer, nayakanti2023wayformer, Rasouli_2021_ICCV}, the manner in which the multimodal data is processed is important. In transformer-based models, besides early or late fusion techniques, hierarchical \cite{nayakanti2023wayformer} and cross-modal fusion methods have been used \cite{rasouli2023pedformer}. In the hierarchical method, different modalities are processed using different attention heads, the output of which are combined via a cross-attention operation. In cross-modal fusion, pair-wise cross-attention units are used to better capture the cross-correlation between different data modalities. However, this approach is computationally expensive as it requires $m(m-1)$ attention modules where $m$ is the number of data modalities. Here, we propose a step-wise hierarchical approach in which different data inputs are gradually fused one at the time by cross-attention modulation. In this way only $m-1$ attention units are needed while the relation between different data types are captured effectively. The outputs of the cross-attention units are concatenated and fed into the encoder transformer.

To model uncertainty, we use a conditional variational autoencoder (CVAE), which generates a latent distribution $p(z|x)$ by learning the similarity to a conditional distribution $q(z|x,y)$ at training time. The combination of the $k$ samples drawn from the latent distribution and the input encodings form the input into the trajectory decoder.
\vspace{-0.1cm}
\subsection{Decoder}
The decoder module has a similar architecture to the encoder, with additional masking operation for prediction. Assuming that the future ego-motion is based on the planned behavior at the time $t$, the encodings are fused with the future ego-motion features before being fed to the decoder. The final output of the decoder is fed into multilayer perceptron layers (MLPs) to infer future trajectories.
\subsubsection{Auxiliary Scaled Prediction} As shown in Sec. \ref{sec:sing-factor}, computing absolute error would bias the error term towards larger scale samples. To balance learning towards smaller scale samples, we add an auxiliary task to the output of the decoder for predicting scaled bounding box coordinates.  
\vspace{-0.1cm}
\subsection{Observation Reconstruction}
One of the key challenges for the egocentric prediction is to separate observed motion resulting from the agent or ego-motion. To achieve better representation, we add a reconstruction task primarily based on ego-motion with a reference to bounding box coordinate at time $t-o+1$ and pedestrian state, i.e., walking or standing. Pedestrian state is used as a signal to indicate whether pedestrian motion is contributing to the observed motion in the image plane. We combine this information via cross-modal attention modulation, where ego-motion and bounding box coordinate serve as a value and the state as a query. The output is fed to the same encoder followed by a separate decoder to generate observed bounding boxes. The reason for reconstructing observation instead of predicting future is twofold: first, observation is shorter and therefore the propagation of error is less, and second, reconstructing observation without a full context can serve as additional supervision signal which is different to future prediction objective.
\vspace{-0.1cm}
\subsection{Objective Function}
We use a separate loss for each regression task, namely $L_{FT}$, $L_{sFT}$ and $L_{ROT}$ which correspond to losses for future trajectories, scaled future trajectories, and reconstructed observed trajectories. An additional KL-divergence (KLD) loss is added for learning the latent distribution. For regression losses we use LogCosh and minimize the losses in a ``best of many'' fashion, e.g., $min_{k}\sum  f(y^{t:t+\tau}, \hat{Y}_k^{t:t+\tau})$. The final loss is a weighted combination of losses as follows:
\vspace{-0.1cm}
$$ L = L_{FT} + \alpha L_{sFT} + \beta L_{ROT} + \gamma KLD, $$

\noindent where weights $\alpha$, $\beta$, and $\gamma$ are set empirically.

\section{Evaluation}

\textbf{Implementation.} We set all input embedding layers to 64 and rest to 128. CVAE MLPs are two layers with 256 and 128 respectively. We use two attention heads for all attention modules in CAs and transformers. Loss weights are set empirically to 10, 2, and 0.1 for $\alpha$, $\beta$, and $\gamma$ respectively. For adjustment, we use width/height ratio of $0.34$. The model is trained for 100 epochs, with a batch size of $128$, learning rate of $4\times10^{-4}$, and Adam optimizer. We refer to our model as \textbf{ENCORE} (\textbf{E}goce\textbf{N}tric predi\textbf{C}ti\textbf{O}n with \textbf{RE}construction). 

\subsection{Scenario-based Analysis}
\begin{table*}[]\vspace{0.2cm}
\caption{Comparison to SOTA on single factor scenarios using $B_{MSE}$/$sB_{MSE}$ metrics. Improvements at the end are computed with respect to PedFormer.}
\label{tbl:single_bmse}
\resizebox{\textwidth}{!}{%
\begin{tabular}{l|ccccccccc|cccccc}
\textbf{Scenario} &
  \multicolumn{7}{c|}{Pedestrian Scale (pixels)} &
  \multicolumn{2}{c|}{Pedestrian State} &
  \multicolumn{6}{c}{Ego-speed (km/h)} \\ \hline
\textbf{Sub-scenario} &
  \multicolumn{1}{c|}{\textbf{0-50}} &
  \multicolumn{1}{c|}{\textbf{50-80}} &
  \multicolumn{1}{c|}{\textbf{80-100}} &
  \multicolumn{1}{c|}{\textbf{100-150}} &
  \multicolumn{1}{c|}{\textbf{150-200}} &
  \multicolumn{1}{c|}{\textbf{200-300}} &
  \multicolumn{1}{c|}{\textbf{300+}} &
  \multicolumn{1}{c|}{\textbf{Walking}} &
  \textbf{Standing} &
  \multicolumn{1}{c|}{\textbf{0}} &
  \multicolumn{1}{c|}{\textbf{0-5}} &
  \multicolumn{1}{c|}{\textbf{5-10}} &
  \multicolumn{1}{c|}{\textbf{10-20}} &
  \multicolumn{1}{c|}{\textbf{20-30}} &
  \textbf{30+} \\ \hline
BiTraP &
  \multicolumn{1}{c|}{134/0.220} &
  \multicolumn{1}{c|}{158/0.097} &
  \multicolumn{1}{c|}{361/0.118} &
  \multicolumn{1}{c|}{538/0.088} &
  \multicolumn{1}{c|}{661/0.054} &
  \multicolumn{1}{c|}{852/0.036} &
  \multicolumn{1}{c|}{1633/0.029} &
  \multicolumn{1}{c|}{579/0.041} &
  400/0.055 &
  \multicolumn{1}{c|}{269/0.019} &
  \multicolumn{1}{c|}{704/0.063} &
  \multicolumn{1}{c|}{1221/0.111} &
  \multicolumn{1}{c|}{1018/0.167} &
  \multicolumn{1}{c|}{597/0.175} &
  386/0.190 \\ 
PedFormer &
  \multicolumn{1}{c|}{76/0.125} &
  \multicolumn{1}{c|}{72/0.044} &
  \multicolumn{1}{c|}{124/0.041} &
  \multicolumn{1}{c|}{250/0.041} &
  \multicolumn{1}{c|}{295/0.024} &
  \multicolumn{1}{c|}{522/0.022} &
  \multicolumn{1}{c|}{1363/0.024} &
  \multicolumn{1}{c|}{394/0.028} &
  153/0.021 &
  \multicolumn{1}{c|}{225/0.016} &
  \multicolumn{1}{c|}{390/0.035} &
  \multicolumn{1}{c|}{487/0.044} &
  \multicolumn{1}{c|}{467/0.077} &
  \multicolumn{1}{c|}{272/0.080} &
  225/0.111 \\ 
\textbf{ENCORE-D (ours)} &
  \multicolumn{1}{c|}{\textbf{55/0.091}} &
  \multicolumn{1}{c|}{\textbf{58/0.036}} &
  \multicolumn{1}{c|}{\textbf{99/0.032}} &
  \multicolumn{1}{c|}{\textbf{189/0.031}} &
  \multicolumn{1}{c|}{\textbf{257/0.021}} &
  \multicolumn{1}{c|}{\textbf{485/0.020}} &
  \multicolumn{1}{c|}{\textbf{1165/0.020}} &
  \multicolumn{1}{c|}{\textbf{351/0.025}} &
  \textbf{108/0.015} &
  \multicolumn{1}{c|}{\textbf{200/0.014}} &
  \multicolumn{1}{c|}{\textbf{320/0.028}} &
  \multicolumn{1}{c|}{\textbf{377/0.034}} &
  \multicolumn{1}{c|}{\textbf{419/0.069}} &
  \multicolumn{1}{c|}{\textbf{217/0.064}} &
  \textbf{145/0.071} \\ \hline 
\textbf{Improvement } &
  \multicolumn{1}{c|}{$27\%$} &
  \multicolumn{1}{c|}{$20\%$} &
  \multicolumn{1}{c|}{$21\%$} &
  \multicolumn{1}{c|}{$24\%$} &
  \multicolumn{1}{c|}{$13\%$} &
  \multicolumn{1}{c|}{$7\%$} &
  \multicolumn{1}{c|}{$15\%$} &
  \multicolumn{1}{c|}{$11\%$} &
  $29\%$ &
  \multicolumn{1}{c|}{$11\%$} &
  \multicolumn{1}{c|}{$18\%$} &
  \multicolumn{1}{c|}{$23\%$} &
  \multicolumn{1}{c|}{$10\%$} &
  \multicolumn{1}{c|}{$20\%$} &
  $36\%$ \\ 
\end{tabular}
\vspace{-0.5cm}
}\vspace{-0.5cm}
\end{table*}

To be comparable to the previous models, we report the results on the deterministic version of our model (ENCORE-D) in this section. 

\noindent\textbf{Single-factor.} In designing ENCORE, we pursued three objectives: better learning of smaller scale samples,  distinguishing between sources of motion, and modeling ego-motion. As results in Table \ref{tbl:single_bmse} suggest, these objectives have been achieved. We can observe significant improvement in smaller scale scenarios, by up $27\%$. In the case of state, there is a significant drop, $29\%$ in the error of Standing scenarios thanks to the use of explicit state encoding. Lastly, we can see that as the overall speed of the vehicle increases, so does the performance gap between ENCORE and PedFormer, reaching the maximum of $36\%$ at 30+ speed. In the case of last step error, $CF_{MSE}$, even more improvement is observable --- up to $37\%$ and $40\%$ in scale and speed scenarios, respectively. 
\begin{figure}[!t]
\vspace{0.2cm}
\centering
\includegraphics[width=1\columnwidth]{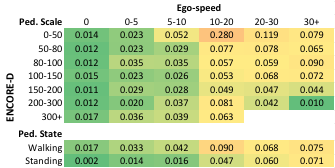}
\caption{Two-factor scenario-based evaluation using $sB_{MSE}$.}
\label{fig:2_factor_encore}
\vspace{-0.5cm}
\end{figure}

\noindent\textbf{Two-factor.} As shown in Figure \ref{fig:2_factor_encore}, ENCORE, performs better across different scenarios with smoother transition across speed dimension. More notable improvements are achieved on smaller scales, 0-100, by up to $37\%$ at low speed 0-5, and $35\%$ at mid-range speed of 10-20. There is also significant improvement in state cases, in particular Standing scenarios. The improvement ratio increases with the ego-speed, peaking at $39\%$.

\begin{table}[]
\caption{Comparison to SOTA on challenging scenarios using $B_{MSE}$/$sB_{MSE}$ metrics.}
\label{tbl:challenge_bmse}
\resizebox{\columnwidth}{!}{%
\begin{tabular}{l|cc|cc|cc}

             & \multicolumn{2}{c|}{\textbf{Pedestrian State}} & \multicolumn{2}{c|}{\textbf{Ego-motion}}    & \multicolumn{2}{c}{\textbf{Ego-action}}    \\ \hline
 &
  \multicolumn{1}{c|}{\textbf{$W_o$-$S_p$}} &
  \textbf{$S_o$-$W_p$} &
  \multicolumn{1}{c|}{\textbf{Constant}} &
  \textbf{Change} &
  \multicolumn{1}{c|}{\textbf{Straight}} &
  \textbf{Turn} \\ \hline
  BiTraP    & \multicolumn{1}{c|}{499/0.059} &	1142/0.079 &		\multicolumn{1}{c|}{430/0.039} &	1899/0.174 &	\multicolumn{1}{c|}{383/0.035}& 3492/0.361 \\
PedFormer    & \multicolumn{1}{c|}{252/0.030}   & 970/0.067   & \multicolumn{1}{c|}{259/0.024} & 924/0.085  & \multicolumn{1}{c|}{248/0.023} & 1422/0.147 \\ 
\textbf{ENCORE-D (ours)} &
  \multicolumn{1}{c|}{\textbf{222/0.026}} &
  \textbf{877/0.061} &
  \multicolumn{1}{c|}{\textbf{225/0.021}} &
  \textbf{707/0.065} &
  \multicolumn{1}{c|}{\textbf{217/0.020}} &
  \textbf{1055/0.109} \\ 
\end{tabular}
\vspace{-0.5cm}
}
\end{table}
\noindent\textbf{Challenging Scenarios.}  A similar pattern of improvement can be observed across different challenging scenarios with more improvement achieved on ego-motion and action by up to $26\%$. This indicates that ENCORE  models ego-motion more effectively compared to past arts. The improvement on pedestrian state is more modest, pointing to the need for more contextual information, e.g., pedestrian pose, signal state, group dynamics, etc. which can be used to deduce future changes in state. 

\subsection{Comparison to SOTA}

\begin{table}[t]
\caption{Comparison to SOTA on the PIE dataset.} \label{tbl:sota_pie}
\resizebox{\columnwidth}{!}{%
\begin{tabular}{l|ccc|c|c|c}
              &\multicolumn{3}{c|}{\textbf{$B_{MSE}$}}                          & \textbf{$C_{MSE}$} & \textbf{$CF_{MSE}$} & \textbf{$sB_{MSE}$} \\ \cline{2-7}
 & \multicolumn{1}{c|}{\textbf{0.5s}} & \multicolumn{1}{c|}{\textbf{1s}} & \multicolumn{1}{c|}{\textbf{1.5s}} & \textbf{1.5s} & \textbf{1.5s} & \textbf{1.5s} \\ \hline
\multicolumn{1}{l|}{B-LSTM \cite{Bhattacharyya_2018_CVPR}}           &\multicolumn{1}{l|}{101} & \multicolumn{1}{c|}{296} & 855  & 811            & 3259            & 0.078        \\ 
\multicolumn{1}{l|}{FOL-X \cite{Yao_2019_ICRA} }           &\multicolumn{1}{c|}{47}  & \multicolumn{1}{c|}{183} & 584  & 546            & 2303            & 0.053        \\ 
\multicolumn{1}{l|}{PIE$_{traj}$ \cite{Rasouli_2019_ICCV}}      & \multicolumn{1}{c|}{58}  & \multicolumn{1}{c|}{200} & 636  & 596            & 2477            & 0.058        \\ 
\multicolumn{1}{l|}{PIE$_{full}$ \cite{Rasouli_2019_ICCV}}       & \multicolumn{1}{c|}{-}    & \multicolumn{1}{c|}{-}    & 556  & 520            & 2162            & 0.051        \\ 
\multicolumn{1}{l|}{BiTraP-D \cite{yao2021bitrap}}         & \multicolumn{1}{c|}{41}  & \multicolumn{1}{c|}{161} & 511  & 481            & 1949            & 0.047        \\ 
\multicolumn{1}{l|}{PEV \cite{Neumann_2021_CVPR}}           &\multicolumn{1}{c|}{42}  & \multicolumn{1}{c|}{153} & 453  & 418            & 1949            & 0.041        \\ 
\multicolumn{1}{l|}{SGNet-ED \cite{wang2022stepwise}}       &\multicolumn{1}{c|}{\textbf{34}}  & \multicolumn{1}{c|}{133} & 442  & 413            & 1761            & 0.040        \\ 
\multicolumn{1}{l|}{BiPed \cite{Rasouli_2021_ICCV}}   & \multicolumn{1}{c|}{{\ul 37}}    & \multicolumn{1}{c|}{119}    & 320  & 291            & 1104            & 0.029        \\ 
\multicolumn{1}{l|}{PedFormer \cite{rasouli2023pedformer}} & \multicolumn{1}{c|}{38}    & \multicolumn{1}{c|}{{\ul 118}}    & {\ul 295}  & {\ul 265}            & {\ul 943}             & {\ul 0.027}        \\ 
\textbf{ENCORE-D (ours)}            &\multicolumn{1}{c|}{{\ul 37}}    & \multicolumn{1}{c|}{\textbf{102}}    & \textbf{251}  & \textbf{222}            & \textbf{805}             & \textbf{0.023}        \\ \hline \hline
\multicolumn{1}{l|}{BiTrap-NP(20) \cite{yao2021bitrap}}    & \multicolumn{1}{c|}{23}  & \multicolumn{1}{c|}{48}  & 102  & 81             & 261             & 0.009        \\ 
\multicolumn{1}{l|}{SGNet-ED(20) \cite{wang2022stepwise}}     & \multicolumn{1}{c|}{{\ul 16}}  & \multicolumn{1}{c|}{39}  & 88   & 66             & 206             & {\ul 0.008}        \\ 
\multicolumn{1}{l|}{ABC+(20) \cite{halawa2022action}}        & \multicolumn{1}{c|}{{\ul 16}}  & \multicolumn{1}{c|}{{\ul 38}}  & {\ul 87}   & {\ul 65}             & {\ul 191}             & {\ul 0.008}        \\ 
\textbf{ENCORE(20) (ours)}      &\multicolumn{1}{c|}{\textbf{15}}    & \multicolumn{1}{c|}{\textbf{33}}    & \textbf{70}   & \textbf{49}             & \textbf{155}             & \textbf{0.006}       
\end{tabular}\vspace{-0.5cm}

}\vspace{-0.5cm}
\end{table}

\begin{table}[t]
\caption{Comparison to SOTA on the JAAD dataset.}\label{tbl:sota_jaad}
\resizebox{\columnwidth}{!}{%
\begin{tabular}{l|ccc|c|c|c}
                 & \multicolumn{3}{c|}{\textbf{$B_{MSE}$}}                              & \textbf{$C_{MSE}$} & \textbf{$CF_{MSE}$} & \textbf{$sB_{MSE}$} \\ \cline{2-7}
 &
  \multicolumn{1}{c|}{\textbf{0.5s}} &
  \multicolumn{1}{c|}{\textbf{1s}} &
  \multicolumn{1}{c|}{\textbf{1.5s}} &
  \multicolumn{1}{c|}{\textbf{1.5s}} &
  \multicolumn{1}{c|}{\textbf{1.5s}} &
  \multicolumn{1}{c}{\textbf{1.5s}} \\ \hline
\multicolumn{1}{l|}{B-LSTM \cite{Bhattacharyya_2018_CVPR}}        & \multicolumn{1}{c|}{159} & \multicolumn{1}{c|}{539}     & 1535 & 1447           & 5617            & 0.122        \\ 
\multicolumn{1}{l|}{FOL-X   \cite{Yao_2019_ICRA}}        & \multicolumn{1}{c|}{147} & \multicolumn{1}{c|}{484}     & 1374 & 1290           & 4924            & 0.110        \\
\multicolumn{1}{l|}{PIE$_{traj}$  \cite{Rasouli_2019_ICCV}}      & \multicolumn{1}{c|}{110} & \multicolumn{1}{c|}{399}     & 1280 & 1183           & 4780            & 0.102        \\ 
\multicolumn{1}{l|}{PIE$_{full}$ \cite{Rasouli_2019_ICCV}}       & \multicolumn{1}{c|}{-}    & \multicolumn{1}{c|}{-}        & 1208 & 1154           & 4717            & 0.096        \\ 
\multicolumn{1}{l|}{BiTraP-D  \cite{yao2021bitrap}}        & \multicolumn{1}{c|}{93}  & \multicolumn{1}{c|}{378}     & 1206 & 1105           & 4565            & 0.096        \\ 
\multicolumn{1}{l|}{PEV \cite{Neumann_2021_CVPR}}            &\multicolumn{1}{c|}{97}  & \multicolumn{1}{c|}{373}     & 1158 & 1042           & 4471            & 0.092        \\ 
\multicolumn{1}{l|}{BiPed \cite{wang2022stepwise}}     & \multicolumn{1}{c|}{85}  & \multicolumn{1}{c|}{362} & 1202 & 1147           & 4759            & 0.096        \\ 
\multicolumn{1}{l|}{PedFormer \cite{rasouli2023pedformer}} & \multicolumn{1}{c|}{93}  & \multicolumn{1}{c|}{364} & 1134 & 1080           & 4364            & 0.090        \\ 
\multicolumn{1}{l|}{SGNet-ED  \cite{wang2022stepwise}}       & \multicolumn{1}{c|}{\textbf{82}}  & \multicolumn{1}{c|}{{\ul 328}}     & {\ul 1049} & {\ul 996}   & {\ul 4076}            & {\ul 0.084}        \\ 
\textbf{ENCORE-NR-D (ours)}         & \multicolumn{1}{c|}{{\ul 83}}    & \multicolumn{1}{c|}{\textbf{319}} & \textbf{980} & \textbf{930}   & \textbf{3766}            & \textbf{0.078}        \\ \hline \hline
\multicolumn{1}{l|}{BiTrap-NP(20) \cite{yao2021bitrap}}    & \multicolumn{1}{c|}{38}  & \multicolumn{1}{c|}{94}      & 222  & 177   & 565             & 0.018        \\ 
\multicolumn{1}{l|}{SGNet-ED(20) \cite{wang2022stepwise}}    & \multicolumn{1}{c|}{{\ul 37}}  & \multicolumn{1}{c|}{{\ul 86}}      & {\ul 197}  & {\ul 146}   & {\ul 443}             & {\ul 0.016}        \\ 
\multicolumn{1}{l|}{ABC+(20) \cite{halawa2022action}}       & \multicolumn{1}{c|}{40}  & \multicolumn{1}{c|}{89}      & \textbf{189}  & \textbf{145}    & \textbf{409}             & \textbf{0.015}        \\
\textbf{ENCORE-NR(20) (ours) }      &  \multicolumn{1}{c|}{\textbf{32}}    & \multicolumn{1}{c|}{\textbf{85}}  & 210  & 167     & 554             & 0.017        
\end{tabular}\vspace{-0.5cm}
}\vspace{-0.4cm}
\end{table}
We evaluate the proposed approach on PIE \cite{Rasouli_2019_ICCV} (introduced earlier) and JAAD \cite{Rasouli_2017_ICCVW}, a commonly used dataset for egocentric trajectory prediction. Note that JAAD does not contain any ego-motion information, hence a modified version of our model with no reconstruction (NR) is evaluated. In addition to the models introduced earlier, we report the results on the following models: FOL-X \cite{Yao_2019_ICRA}, PEV \cite{Neumann_2021_CVPR}, BiPed \cite{Rasouli_2021_ICCV} (from which, similar to PedFormer, we remove the action prediction task), SGNet-ED \cite{wang2022stepwise}, ABC+ \cite{halawa2022action}. As for metrics, we report on $B_{mse}$, $sB_{mse}$, $C_{mse}$, and $CF_{mse}$.

Tables \ref{tbl:sota_pie} and \ref{tbl:sota_jaad} summarize the results of our model on both datasets. On PIE, both versions of our model significantly improve upon the past arts on most metrics, ENCORE-D improves performance by up to $16\%$ compared to PedFormer and ENCORE by up $25\%$ compared to ABC+. On JAAD, improvement of up to $8\%$ is achieved on the deterministic model, while the non-deterministic model improves on two metrics (by up to $13\%$) and achieves comparable results to the past arts on the rest. One reason for a smaller gap is that the majority of JAAD samples are small-scale, therefore the adjustment offers limited improvement and its effect is lowered in non-deterministic modeling.

\begin{table}[t]
\caption{Ablation study results on proposed modules.}\label{tbl:abl_model}
\resizebox{\columnwidth}{!}{%
\begin{tabular}{c|c|c|c|cccc}
\textbf{HSF} & \textbf{sFT} & \textbf{POFT} & \textbf{ROT} & \multicolumn{1}{c}{\textbf{$B_{MSE}$}} & \multicolumn{1}{c}{\textbf{$C_{MSE}$}} & \multicolumn{1}{c}{\textbf{$CF_{MSE}$}} & \textbf{$sB_{MSE}$}          \\ \hline
  &   &   &  & 90 & 66 & 248 & 0.008 \\ \hline
\checkmark &   &   &  & 80 & 58 & 232 & 0.007 \\ \hline
\checkmark & \checkmark &   &  & 77 & 56 & 198 & 0.007 \\ \hline
\checkmark & \checkmark & \checkmark &  & 79 & 56 & 222 & 0.007 \\ \hline
\checkmark   & \checkmark   &     & \checkmark   & \textbf{70}               & \textbf{49}               & \textbf{155}               & \textbf{0.006}
\end{tabular}
\vspace{-0.5cm}
}
\vspace{-0.4cm}
\end{table}
\noindent\textbf{Ablation Study.} We examine the effectiveness of the proposed modules, namely hierarchical step-wise fusion (HSF) (compared to the approach in \cite{rasouli2023pedformer}), scaled future trajectories (sFT), and partial observation for future trajectories (POFT) as an alternative to the proposed reconstruction of observation trajectories (ROT). As shown in Table \ref{tbl:abl_model}, the introduction of each module has a positive impact mainly on the average metrics. Reconstruction module, on the other hand, improves upon on all metric, with more significant impact on final error metric. This is due to the regularization effect that reconstruction has to minimize error propagation, something that is often achieved with explicit goal setting in the  past arts \cite{yao2021bitrap,rasouli2023pedformer,wang2022stepwise}. It should be noted that, as shown in the table, predicting future trajectories based on partial observable data, instead of reconstructing input, does not provide any benefits as was anticipated. 

\section{Conclusion}
We proposed a new approach to evaluating egocentric pedestrian trajectory prediction models based on various contextual factors extracted from data. We highlighted different challenges posed to the models stemming from diverse set of factors belonging to pedestrians and ego-agent. Based on our findings, we proposed a novel model, ENCORE, which achieves state-of-the-art performance by significantly improving upon the past arts. We conducted scenario-based analysis using the newly proposed model and ablation studies and showed how effective our approach in resolving common challenges in egocentric prediction.
\bibliographystyle{IEEEtran}

\section{Supplementary Materials}

\subsection{Adjustment Effect on Metric Value}
\begin{table}[h]
\caption{The effect of bounding box adjustment in calculation of $sB_{MSE}$ metric. Results are reported for PedFormer model on different pedestrian scale scenarios.}
\label{tbl:adjust}
\resizebox{\columnwidth}{!}{%
\begin{tabular}{l|c|c|c|c|c|c|c}

       & \textbf{0-50} & \textbf{50-80} & \textbf{80-100} & \textbf{100-150} & \textbf{150-200} & \textbf{200-300} & \textbf{300+} \\ \hline
Not adjusted & 0.1339        & 0.0480         & 0.0440          & 0.0448           & 0.0260           & 0.0232           & 0.0254        \\ \hline
Adjusted    & 0.1249        & 0.0445         & 0.0407          & 0.0410           & 0.0243           & 0.0218           & 0.0239        \\ 
\end{tabular}%
}
\end{table}
As discussed in the paper, bounding boxes in the scenes are often truncated due to occlusions or on the scene boundaries. This can lead to miscalculation of the scaled error metrics. Hence, the bounding boxes' aspect ratios are adjusted in order to compute the scaled metrics. The changes in metric values with and without adjustment for a single-factor scenario are shown in Table \ref{tbl:adjust}.

\subsection{Single Factor Scenarios}

\begin{table*}[th]
\caption{Comparison to SOTA on single-factor scenarios using $B_{MSE}$/$sB_{MSE}$ metrics. Improvements at the end are computed with respect to PedFormer.}
\label{tbl:single_bmse}
\resizebox{\textwidth}{!}{%
\begin{tabular}{l|ccccccccc|cccccc}

 &
  \multicolumn{9}{c|}{Pedestrian} &
  \multicolumn{6}{c}{Ego} \\ \hline
\textbf{Scenario} &
  \multicolumn{7}{c|}{Scale (pixels)} &
  \multicolumn{2}{c|}{State} &
  \multicolumn{6}{c}{Speed (km/h)} \\ \hline
\textbf{Sub-scenario} &
  \multicolumn{1}{c|}{\textbf{0-50}} &
  \multicolumn{1}{c|}{\textbf{50-80}} &
  \multicolumn{1}{c|}{\textbf{80-100}} &
  \multicolumn{1}{c|}{\textbf{100-150}} &
  \multicolumn{1}{c|}{\textbf{150-200}} &
  \multicolumn{1}{c|}{\textbf{200-300}} &
  \multicolumn{1}{c|}{\textbf{300+}} &
  \multicolumn{1}{c|}{\textbf{Walking}} &
  \textbf{Standing} &
  \multicolumn{1}{c|}{\textbf{0}} &
  \multicolumn{1}{c|}{\textbf{0-5}} &
  \multicolumn{1}{c|}{\textbf{5-10}} &
  \multicolumn{1}{c|}{\textbf{10-20}} &
  \multicolumn{1}{c|}{\textbf{20-30}} &
  \textbf{30+} \\ \hline
PIE$_{traj}$ &
  \multicolumn{1}{c|}{187/0.308} &
  \multicolumn{1}{c|}{170/0.105} &
  \multicolumn{1}{c|}{373/0.122} &
  \multicolumn{1}{c|}{595/0.098} &
  \multicolumn{1}{c|}{709/0.058} &
  \multicolumn{1}{c|}{980/0.041} &
  \multicolumn{1}{c|}{2064/0.036} &
  \multicolumn{1}{c|}{677/0.048} &
  441/0.060 &
  \multicolumn{1}{c|}{284/0.020} &
  \multicolumn{1}{c|}{800/0.071} &
  \multicolumn{1}{c|}{1495/0.136} &
  \multicolumn{1}{c|}{1153/0.189} &
  \multicolumn{1}{c|}{709/0.208} &
  649/0.320 \\ \hline
B-LSTM &
  \multicolumn{1}{c|}{225/0.371} &
  \multicolumn{1}{c|}{215/0.132} &
  \multicolumn{1}{c|}{464/0.152} &
  \multicolumn{1}{c|}{840/0.138} &
  \multicolumn{1}{c|}{886/0.073} &
  \multicolumn{1}{c|}{1170/0.049} &
  \multicolumn{1}{c|}{2492/0.044} &
  \multicolumn{1}{c|}{833/0.059} &
  582/0.080 &
  \multicolumn{1}{c|}{367/0.026} &
  \multicolumn{1}{c|}{884/0.079} &
  \multicolumn{1}{c|}{1849/0.168} &
  \multicolumn{1}{c|}{1405/0.231} &
  \multicolumn{1}{c|}{954/0.280} &
  969/0.477 \\ \hline
PIE$_{full}$ &
  \multicolumn{1}{c|}{195/0.322} &
  \multicolumn{1}{c|}{163/0.101} &
  \multicolumn{1}{c|}{313/0.102} &
  \multicolumn{1}{c|}{447/0.074} &
  \multicolumn{1}{c|}{513/0.042} &
  \multicolumn{1}{c|}{628/0.026} &
  \multicolumn{1}{c|}{1838/0.032} &
  \multicolumn{1}{c|}{559/0.040} &
  337/0.046 &
  \multicolumn{1}{c|}{259/0.019} &
  \multicolumn{1}{c|}{659/0.059} &
  \multicolumn{1}{c|}{939/0.085} &
  \multicolumn{1}{c|}{976/0.160} &
  \multicolumn{1}{c|}{563/0.165} &
  317/0.156 \\ \hline
BiTraP &
  \multicolumn{1}{c|}{134/0.220} &
  \multicolumn{1}{c|}{158/0.097} &
  \multicolumn{1}{c|}{361/0.118} &
  \multicolumn{1}{c|}{538/0.088} &
  \multicolumn{1}{c|}{661/0.054} &
  \multicolumn{1}{c|}{852/0.036} &
  \multicolumn{1}{c|}{1633/0.029} &
  \multicolumn{1}{c|}{579/0.041} &
  400/0.055 &
  \multicolumn{1}{c|}{269/0.019} &
  \multicolumn{1}{c|}{704/0.063} &
  \multicolumn{1}{c|}{1221/0.111} &
  \multicolumn{1}{c|}{1018/0.167} &
  \multicolumn{1}{c|}{597/0.175} &
  386/0.190 \\ \hline
BiPed &
  \multicolumn{1}{c|}{96/0.158} &
  \multicolumn{1}{c|}{82/0.051} &
  \multicolumn{1}{c|}{138/0.045} &
  \multicolumn{1}{c|}{246/0.040} &
  \multicolumn{1}{c|}{323/0.027} &
  \multicolumn{1}{c|}{541/0.023} &
  \multicolumn{1}{c|}{1520/0.027} &
  \multicolumn{1}{c|}{429/0.030} &
  165/0.023 &
  \multicolumn{1}{c|}{233/0.017} &
  \multicolumn{1}{c|}{444/0.039} &
  \multicolumn{1}{c|}{551/0.050} &
  \multicolumn{1}{c|}{522/0.086} &
  \multicolumn{1}{c|}{272/0.080} &
  282/0.139 \\ \hline
PedFormer &
  \multicolumn{1}{c|}{76/0.125} &
  \multicolumn{1}{c|}{72/0.044} &
  \multicolumn{1}{c|}{124/0.041} &
  \multicolumn{1}{c|}{250/0.041} &
  \multicolumn{1}{c|}{295/0.024} &
  \multicolumn{1}{c|}{522/0.022} &
  \multicolumn{1}{c|}{1363/0.024} &
  \multicolumn{1}{c|}{394/0.028} &
  153/0.021 &
  \multicolumn{1}{c|}{225/0.016} &
  \multicolumn{1}{c|}{390/0.035} &
  \multicolumn{1}{c|}{487/0.044} &
  \multicolumn{1}{c|}{467/0.077} &
  \multicolumn{1}{c|}{272/0.080} &
  225/0.111 \\ \hline
\textbf{ENCORE-D} &
  \multicolumn{1}{c|}{\textbf{55/0.091}} &
  \multicolumn{1}{c|}{\textbf{58/0.036}} &
  \multicolumn{1}{c|}{\textbf{99/0.032}} &
  \multicolumn{1}{c|}{\textbf{189/0.031}} &
  \multicolumn{1}{c|}{\textbf{257/0.021}} &
  \multicolumn{1}{c|}{\textbf{485/0.020}} &
  \multicolumn{1}{c|}{\textbf{1165/0.020}} &
  \multicolumn{1}{c|}{\textbf{351/0.025}} &
  \textbf{108/0.015} &
  \multicolumn{1}{c|}{\textbf{200/0.014}} &
  \multicolumn{1}{c|}{\textbf{320/0.028}} &
  \multicolumn{1}{c|}{\textbf{377/0.034}} &
  \multicolumn{1}{c|}{\textbf{419/0.069}} &
  \multicolumn{1}{c|}{\textbf{217/0.064}} &
  \textbf{145/0.071} \\ \hline
\textbf{Improvement } &
  \multicolumn{1}{c|}{$27\%$} &
  \multicolumn{1}{c|}{$20\%$} &
  \multicolumn{1}{c|}{$21\%$} &
  \multicolumn{1}{c|}{$24\%$} &
  \multicolumn{1}{c|}{$13\%$} &
  \multicolumn{1}{c|}{$7\%$} &
  \multicolumn{1}{c|}{$15\%$} &
  \multicolumn{1}{c|}{$11\%$} &
  $29\%$ &
  \multicolumn{1}{c|}{$11\%$} &
  \multicolumn{1}{c|}{$18\%$} &
  \multicolumn{1}{c|}{$23\%$} &
  \multicolumn{1}{c|}{$10\%$} &
  \multicolumn{1}{c|}{$20\%$} &
  $36\%$ \\ 
\end{tabular}%
}
\end{table*}

\begin{table*}[th]
\caption{Comparison to SOTA on single-factor scenarios using $C_{MSE}$/$sC_{MSE}$ metrics. Improvements at the end are computed with respect to PedFormer.}
\label{tbl:single_cmse}
\resizebox{\textwidth}{!}{%
\begin{tabular}{l|ccccccccc|cccccc}

 &
  \multicolumn{9}{c|}{Pedestrian} &
  \multicolumn{6}{c}{Ego} \\ \hline
\textbf{Scenario} &
  \multicolumn{7}{c|}{Scale (pixels)} &
  \multicolumn{2}{c|}{State} &
  \multicolumn{6}{c}{Speed (km/h)} \\ \hline
\textbf{Sub-scenario} &
  \multicolumn{1}{c|}{\textbf{0-50}} &
  \multicolumn{1}{c|}{\textbf{50-80}} &
  \multicolumn{1}{c|}{\textbf{80-100}} &
  \multicolumn{1}{c|}{\textbf{100-150}} &
  \multicolumn{1}{c|}{\textbf{150-200}} &
  \multicolumn{1}{c|}{\textbf{200-300}} &
  \multicolumn{1}{c|}{\textbf{300+}} &
  \multicolumn{1}{c|}{\textbf{Walking}} &
  \textbf{Standing} &
  \multicolumn{1}{c|}{\textbf{0}} &
  \multicolumn{1}{c|}{\textbf{0-5}} &
  \multicolumn{1}{c|}{\textbf{5-10}} &
  \multicolumn{1}{c|}{\textbf{10-20}} &
  \multicolumn{1}{c|}{\textbf{20-30}} &
  \textbf{30+} \\ \hline
PIE$_{traj}$ &
  \multicolumn{1}{c|}{185/0.305} &
  \multicolumn{1}{c|}{166/0.103} &
  \multicolumn{1}{c|}{366/0.120} &
  \multicolumn{1}{c|}{573/0.094} &
  \multicolumn{1}{c|}{671/0.055} &
  \multicolumn{1}{c|}{901/0.038} &
  \multicolumn{1}{c|}{1899/0.033} &
  \multicolumn{1}{c|}{633/0.045} &
  424/0.058 &
  \multicolumn{1}{c|}{257/0.018} &
  \multicolumn{1}{c|}{757/0.067} &
  \multicolumn{1}{c|}{1431/0.130} &
  \multicolumn{1}{c|}{1108/0.182} &
  \multicolumn{1}{c|}{679/0.199} &
  625/0.308 \\ \hline
B-LSTM &
  \multicolumn{1}{c|}{222/0.367} &
  \multicolumn{1}{c|}{210/0.130} &
  \multicolumn{1}{c|}{456/0.149} &
  \multicolumn{1}{c|}{815/0.134} &
  \multicolumn{1}{c|}{841/0.069} &
  \multicolumn{1}{c|}{1076/0.045} &
  \multicolumn{1}{c|}{2303/0.040} &
  \multicolumn{1}{c|}{781/0.055} &
  561/0.077 &
  \multicolumn{1}{c|}{335/0.024} &
  \multicolumn{1}{c|}{838/0.074} &
  \multicolumn{1}{c|}{1783/0.162} &
  \multicolumn{1}{c|}{1351/0.222} &
  \multicolumn{1}{c|}{916/0.269} &
  938/0.462 \\ \hline
PIE$_{full}$ &
  \multicolumn{1}{c|}{194/0.319} &
  \multicolumn{1}{c|}{160/0.099} &
  \multicolumn{1}{c|}{307/0.100} &
  \multicolumn{1}{c|}{430/0.071} &
  \multicolumn{1}{c|}{482/0.040} &
  \multicolumn{1}{c|}{564/0.024} &
  \multicolumn{1}{c|}{1685/0.030} &
  \multicolumn{1}{c|}{519/0.037} &
  325/0.045 &
  \multicolumn{1}{c|}{235/0.017} &
  \multicolumn{1}{c|}{619/0.055} &
  \multicolumn{1}{c|}{884/0.080} &
  \multicolumn{1}{c|}{942/0.155} &
  \multicolumn{1}{c|}{543/0.159} &
  302/0.148 \\ \hline
BiTraP &
  \multicolumn{1}{c|}{132/0.218} &
  \multicolumn{1}{c|}{154/0.095} &
  \multicolumn{1}{c|}{354/0.116} &
  \multicolumn{1}{c|}{517/0.085} &
  \multicolumn{1}{c|}{624/0.051} &
  \multicolumn{1}{c|}{782/0.033} &
  \multicolumn{1}{c|}{1487/0.026} &
  \multicolumn{1}{c|}{539/0.038} &
  384/0.053 &
  \multicolumn{1}{c|}{244/0.017} &
  \multicolumn{1}{c|}{665/0.059} &
  \multicolumn{1}{c|}{1161/0.105} &
  \multicolumn{1}{c|}{975/0.160} &
  \multicolumn{1}{c|}{576/0.169} &
  368/0.181 \\ \hline
BiPed &
  \multicolumn{1}{c|}{94/0.154} &
  \multicolumn{1}{c|}{79/0.049} &
  \multicolumn{1}{c|}{132/0.043} &
  \multicolumn{1}{c|}{228/0.038} &
  \multicolumn{1}{c|}{292/0.024} &
  \multicolumn{1}{c|}{474/0.020} &
  \multicolumn{1}{c|}{1364/0.024} &
  \multicolumn{1}{c|}{388/0.027} &
  153/0.021 &
  \multicolumn{1}{c|}{208/0.015} &
  \multicolumn{1}{c|}{405/0.036} &
  \multicolumn{1}{c|}{493/0.045} &
  \multicolumn{1}{c|}{487/0.080} &
  \multicolumn{1}{c|}{253/0.074} &
  266/0.131 \\ \hline
PedFormer &
  \multicolumn{1}{c|}{73/0.121} &
  \multicolumn{1}{c|}{68/0.042} &
  \multicolumn{1}{c|}{118/0.039} &
  \multicolumn{1}{c|}{231/0.038} &
  \multicolumn{1}{c|}{263/0.022} &
  \multicolumn{1}{c|}{453/0.019} &
  \multicolumn{1}{c|}{1201/0.021} &
  \multicolumn{1}{c|}{350/0.025} &
  142/0.019 &
  \multicolumn{1}{c|}{197/0.014} &
  \multicolumn{1}{c|}{349/0.031} &
  \multicolumn{1}{c|}{430/0.039} &
  \multicolumn{1}{c|}{432/0.071} &
  \multicolumn{1}{c|}{254/0.075} &
  213/0.105 \\ \hline
\textbf{ENCORE-D} &
  \multicolumn{1}{c|}{\textbf{55/0.090}} &
  \multicolumn{1}{c|}{\textbf{55/0.034}} &
  \multicolumn{1}{c|}{\textbf{94/0.031}} &
  \multicolumn{1}{c|}{\textbf{175/0.029}} &
  \multicolumn{1}{c|}{\textbf{227/0.019}} &
  \multicolumn{1}{c|}{\textbf{418/0.017}} &
  \multicolumn{1}{c|}{\textbf{1001/0.018}} &
  \multicolumn{1}{c|}{\textbf{308/0.022}} &
  \textbf{99/0.014} &
  \multicolumn{1}{c|}{\textbf{174/0.012}} &
  \multicolumn{1}{c|}{\textbf{280/0.025}} &
  \multicolumn{1}{c|}{\textbf{321/0.029}} &
  \multicolumn{1}{c|}{\textbf{385/0.063}} &
  \multicolumn{1}{c|}{\textbf{203/0.060}} &
  \textbf{138/0.068} \\ \hline
\textbf{Improvement } &
  \multicolumn{1}{c|}{$25\%$} &
  \multicolumn{1}{c|}{$19\%$} &
  \multicolumn{1}{c|}{$20\%$} &
  \multicolumn{1}{c|}{$25\%$} &
  \multicolumn{1}{c|}{$14\%$} &
  \multicolumn{1}{c|}{$8\%$} &
  \multicolumn{1}{c|}{$17\%$} &
  \multicolumn{1}{c|}{$12\%$} &
  $30\%$ &
  \multicolumn{1}{c|}{$12\%$} &
  \multicolumn{1}{c|}{$20\%$} &
  \multicolumn{1}{c|}{$25\%$} &
  \multicolumn{1}{c|}{$11\%$} &
  \multicolumn{1}{c|}{$20\%$} &
  $35\%$ \\
\end{tabular}%
}
\end{table*}

\begin{table*}[th]
\caption{Comparison to SOTA on single-factor scenarios using $CF_{MSE}$/$sCF_{MSE}$ metrics. Improvements at the end are computed with respect to PedFormer.}
\label{tbl:single_cfmse}
\resizebox{\textwidth}{!}{%
\begin{tabular}{l|ccccccccc|cccccc}
 &
  \multicolumn{9}{c|}{Pedestrian} &
  \multicolumn{6}{c}{Ego} \\ \hline
\textbf{Scenario} &
  \multicolumn{7}{c|}{Scale (pixels)} &
  \multicolumn{2}{c|}{State} &
  \multicolumn{6}{c}{Speed (km/h)} \\ \hline
\textbf{Sub-scenario} &
  \multicolumn{1}{c|}{\textbf{0-50}} &
  \multicolumn{1}{c|}{\textbf{50-80}} &
  \multicolumn{1}{c|}{\textbf{80-100}} &
  \multicolumn{1}{c|}{\textbf{100-150}} &
  \multicolumn{1}{c|}{\textbf{150-200}} &
  \multicolumn{1}{c|}{\textbf{200-300}} &
  \multicolumn{1}{c|}{\textbf{300+}} &
  \multicolumn{1}{c|}{\textbf{Walking}} &
  \textbf{Standing} &
  \multicolumn{1}{c|}{\textbf{0}} &
  \multicolumn{1}{c|}{\textbf{0-5}} &
  \multicolumn{1}{c|}{\textbf{5-10}} &
  \multicolumn{1}{c|}{\textbf{10-20}} &
  \multicolumn{1}{c|}{\textbf{20-30}} &
  \textbf{30+} \\ \hline
PIE$_{traj}$ &
  \multicolumn{1}{c|}{642/0.906} &
  \multicolumn{1}{c|}{694/0.361} &
  \multicolumn{1}{c|}{1711/0.456} &
  \multicolumn{1}{c|}{2930/0.368} &
  \multicolumn{1}{c|}{3150/0.219} &
  \multicolumn{1}{c|}{3858/0.141} &
  \multicolumn{1}{c|}{7061/0.119} &
  \multicolumn{1}{c|}{2551/0.168} &
  2041/0.230 &
  \multicolumn{1}{c|}{873/0.062} &
  \multicolumn{1}{c|}{3289/0.256} &
  \multicolumn{1}{c|}{6232/0.424} &
  \multicolumn{1}{c|}{4951/0.498} &
  \multicolumn{1}{c|}{3468/0.527} &
  3734/0.837 \\ \hline
B-LSTM &
  \multicolumn{1}{c|}{701/0.989} &
  \multicolumn{1}{c|}{882/0.459} &
  \multicolumn{1}{c|}{2224/0.593} &
  \multicolumn{1}{c|}{4190/0.526} &
  \multicolumn{1}{c|}{3937/0.274} &
  \multicolumn{1}{c|}{4533/0.166} &
  \multicolumn{1}{c|}{8809/0.149} &
  \multicolumn{1}{c|}{3116/0.205} &
  2784/0.313 &
  \multicolumn{1}{c|}{1130/0.080} &
  \multicolumn{1}{c|}{3611/0.281} &
  \multicolumn{1}{c|}{7554/0.513} &
  \multicolumn{1}{c|}{5942/0.598} &
  \multicolumn{1}{c|}{5051/0.767} &
  5944/1.333 \\ \hline
PIE$_{full}$ &
  \multicolumn{1}{c|}{623/0.879} &
  \multicolumn{1}{c|}{678/0.353} &
  \multicolumn{1}{c|}{1420/0.379} &
  \multicolumn{1}{c|}{2082/0.261} &
  \multicolumn{1}{c|}{2077/0.145} &
  \multicolumn{1}{c|}{2194/0.080} &
  \multicolumn{1}{c|}{6115/0.103} &
  \multicolumn{1}{c|}{1974/0.130} &
  1467/0.165 &
  \multicolumn{1}{c|}{779/0.055} &
  \multicolumn{1}{c|}{2640/0.206} &
  \multicolumn{1}{c|}{3905/0.265} &
  \multicolumn{1}{c|}{3971/0.400} &
  \multicolumn{1}{c|}{2500/0.380} &
  1536/0.344 \\ \hline
BiTraP &
  \multicolumn{1}{c|}{459/0.648} &
  \multicolumn{1}{c|}{634/0.330} &
  \multicolumn{1}{c|}{1637/0.437} &
  \multicolumn{1}{c|}{2395/0.301} &
  \multicolumn{1}{c|}{2789/0.194} &
  \multicolumn{1}{c|}{3124/0.115} &
  \multicolumn{1}{c|}{5174/0.088} &
  \multicolumn{1}{c|}{2087/0.137} &
  1713/0.193 &
  \multicolumn{1}{c|}{821/0.058} &
  \multicolumn{1}{c|}{2873/0.224} &
  \multicolumn{1}{c|}{4664/0.317} &
  \multicolumn{1}{c|}{4203/0.423} &
  \multicolumn{1}{c|}{2761/0.419} &
  2009/0.450 \\ \hline
BiPed &
  \multicolumn{1}{c|}{309/0.436} &
  \multicolumn{1}{c|}{312/0.162} &
  \multicolumn{1}{c|}{542/0.145} &
  \multicolumn{1}{c|}{976/0.123} &
  \multicolumn{1}{c|}{1138/0.079} &
  \multicolumn{1}{c|}{1817/0.067} &
  \multicolumn{1}{c|}{4868/0.082} &
  \multicolumn{1}{c|}{1417/0.093} &
  638/0.072 &
  \multicolumn{1}{c|}{697/0.050} &
  \multicolumn{1}{c|}{1555/0.121} &
  \multicolumn{1}{c|}{1982/0.135} &
  \multicolumn{1}{c|}{1975/0.199} &
  \multicolumn{1}{c|}{1090/0.165} &
  1289/0.289 \\ \hline
PedFormer &
  \multicolumn{1}{c|}{225/0.318} &
  \multicolumn{1}{c|}{239/0.124} &
  \multicolumn{1}{c|}{451/0.120} &
  \multicolumn{1}{c|}{900/0.113} &
  \multicolumn{1}{c|}{967/0.067} &
  \multicolumn{1}{c|}{1628/0.060} &
  \multicolumn{1}{c|}{4089/0.069} &
  \multicolumn{1}{c|}{1208/0.079} &
  535/0.060 &
  \multicolumn{1}{c|}{642/0.046} &
  \multicolumn{1}{c|}{1230/0.096} &
  \multicolumn{1}{c|}{1540/0.105} &
  \multicolumn{1}{c|}{1559/0.157} &
  \multicolumn{1}{c|}{1124/0.171} &
  968/0.217 \\ \hline
\textbf{ENCORE-D} &
  \multicolumn{1}{c|}{\textbf{142/0.201}} &
  \multicolumn{1}{c|}{\textbf{184/0.096}} &
  \multicolumn{1}{c|}{\textbf{360/0.096}} &
  \multicolumn{1}{c|}{\textbf{737/0.093}} &
  \multicolumn{1}{c|}{\textbf{836/0.058}} &
  \multicolumn{1}{c|}{\textbf{1539/0.056}} &
  \multicolumn{1}{c|}{\textbf{3496/0.059}} &
  \multicolumn{1}{c|}{\textbf{1074/0.071}} &
  \textbf{390/0.044} &
  \multicolumn{1}{c|}{\textbf{571/0.041}} &
  \multicolumn{1}{c|}{\textbf{999/0.078}} &
  \multicolumn{1}{c|}{\textbf{1307/0.089}} &
  \multicolumn{1}{c|}{\textbf{1513/0.152}} &
  \multicolumn{1}{c|}{\textbf{831/0.126}} &
  \textbf{580/0.130} \\ \hline
\textbf{Improvement } &
  \multicolumn{1}{c|}{$37\%$} &
  \multicolumn{1}{c|}{$23\%$} &
  \multicolumn{1}{c|}{$20\%$} &
  \multicolumn{1}{c|}{$18\%$} &
  \multicolumn{1}{c|}{$14\%$} &
  \multicolumn{1}{c|}{$5\%$} &
  \multicolumn{1}{c|}{$14\%$} &
  \multicolumn{1}{c|}{$11\%$} &
  $27\%$ &
  \multicolumn{1}{c|}{$11\%$} &
  \multicolumn{1}{c|}{$19\%$} &
  \multicolumn{1}{c|}{$15\%$} &
  \multicolumn{1}{c|}{$3\%$} &
  \multicolumn{1}{c|}{$26\%$} &
  $40\%$ \\ 
\end{tabular}
}
\end{table*}

In this section we report a full comparison of the proposed \textbf{ENCORE} model against SOTA models, including \textbf{PIE$_{traj}$} \cite{Rasouli_2019_ICCV}, \textbf{PIE$_{full}$}, \textbf{B-LSTM} \cite{Bhattacharyya_2018_CVPR}, \textbf{BiTraP}  \cite{yao2021bitrap}, \textbf{BiPed} \cite{Rasouli_2021_ICCV}, and \textbf{PedFormer} \cite{rasouli2023pedformer}. As before, we evaluate using the deterministic version of the proposed model, ENCORE-D.

The results are reported per metric in Tables \ref{tbl:single_bmse} \ref{tbl:single_cmse}, and \ref{tbl:single_cfmse}. As expected, ENCORE-D achieves significant improvement across all metrics and scenarios. The degree of improvement, however, varies. Overall, in scale scenarios, we can observe more improvement on smaller scale cases 0-150 pixels where improvements of up to $27\%$ on average error ($B_{mse}$) and $37\%$ on final step error, $CF_{MSE}$ are achieved. The improvements are due to the effect of the proposed auxiliary task to predict scaled bounding boxes which forces the model to pay more attention to smaller scale boxes. The larger improvement on final step error can be attributed to the regularization impact of the reconstruction module.

Another major improvement can be seen in the case of the Standing scenarios where improvement of $30\%$ is achieved compared to $12\%$ in the case of walking scenarios. This is thanks to the use of explicit signals regarding pedestrian state telling the model that more attention should be given to ego-motion when the pedestrian is immobile.

Lastly, our proposed model, ENCORE-D, results in better modeling of ego-motion where the effect becomes more apparent as the speed of the vehicle increases. In the case of the highest speed category, 30+,  $40\%$ improvement is achieved. 

\subsection{Two Factor Scenarios}
\begin{figure}[!t]
\vspace{0.2cm}
\centering
\includegraphics[width=1\columnwidth]{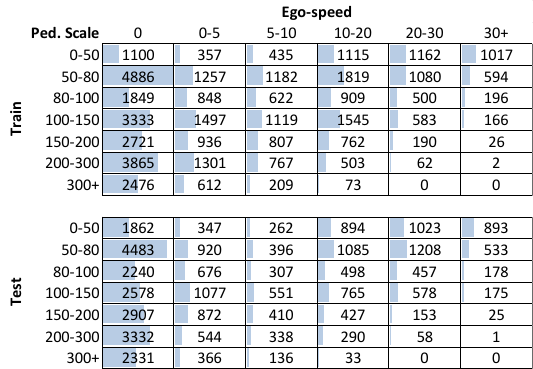}
\caption{Train and test data counts for two-factor scenarios.}
\label{fig:2_factor_data_speed_scale}
\end{figure}

\begin{figure}[!t]
\vspace{0.2cm}
\centering
\includegraphics[width=1\columnwidth]{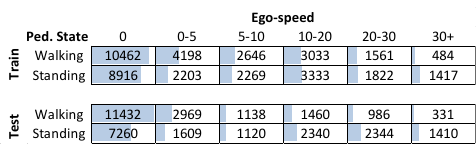}
\caption{Train and test data counts for two-factor scenarios.}
\label{fig:2_factor_data_speed_state}
\end{figure}

Besides challenges associated with the scenarios, the number of samples used for training the models plays a vital role in the overall performance. To shed more light on per-scenario performance of the models, data counts for two-factor scenarios, both for train and test subsets are shown in Figures \ref{fig:2_factor_data_speed_scale} and \ref{fig:2_factor_data_speed_state}. It can be seen that in some cases the number of samples become very small, e.g. in the case of higher speed and larger scale. Therefore, extraction of scenarios with additional factors (e.g. for three-factor analysis) can result in either non-existing samples or scenarios with very few samples.

\begin{figure*}[!t]
\vspace{0.2cm}
\centering
\includegraphics[width=1\textwidth]{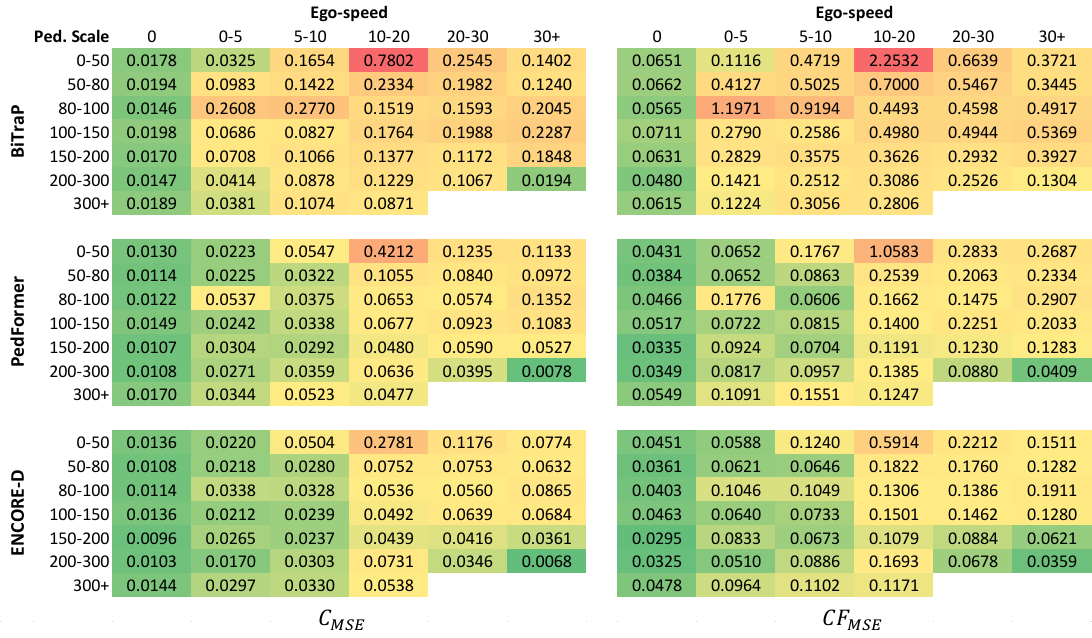}
\caption{Two-factor scenario-based analysis using center metrics.}
\label{fig:2_factor_sup_scale_speed}
\end{figure*}

\begin{figure*}[!t]
\vspace{0.2cm}
\centering
\includegraphics[width=1\textwidth]{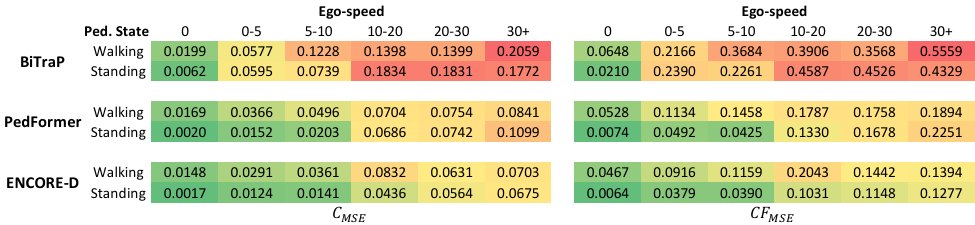}
\caption{Two-factor scenario-based analysis using center metrics.}
\label{fig:2_factor_sup_speed_state}
\vspace{-0.5cm}
\end{figure*}

We additionally report on the performance of the models (as in the paper) for two-factor scenarios using two additional metrics. As shown in Figures \ref{fig:2_factor_sup_scale_speed} and \ref{fig:2_factor_sup_speed_state}, the proposed model, ENCORE-D, clearly stands out, especially in more challenging cases, such as smaller scale prediction or higher ego-speed. Once again the improvement is more apparent in the case of pedestrian state vs ego-speed scenarios.

\subsection{Challenging Scenarios}

\begin{table}[h]
\caption{Comparison to SOTA on challenging scenarios using $B_{MSE}$/$sB_{MSE}$ metrics.}
\label{tbl:challenge_bmse}
\resizebox{\columnwidth}{!}{%
\begin{tabular}{l|cc|cc|cc}
             & \multicolumn{2}{c|}{\textbf{Pedestrian State}} & \multicolumn{2}{c|}{\textbf{Ego-motion}}    & \multicolumn{2}{c}{\textbf{Ego-action}}    \\ \hline
 &
  \multicolumn{1}{c|}{\textbf{$W_o$-$S_p$}} &
  \textbf{$S_o$-$W_p$} &
  \multicolumn{1}{c|}{\textbf{Constant}} &
  \textbf{Change} &
  \multicolumn{1}{c|}{\textbf{Straight}} &
  \textbf{Turn} \\ \hline
PIE$_{traj}$ & \multicolumn{1}{c|}{536/0.064}   & 1553/0.108  & \multicolumn{1}{c|}{474/0.043} & 2576/0.236 & \multicolumn{1}{c|}{436/0.040} & 4166/0.430 \\ \hline
B-LSTM       & \multicolumn{1}{c|}{513/0.061}   & 1847/0.128  & \multicolumn{1}{c|}{593/0.054} & 3213/0.295 & \multicolumn{1}{c|}{542/0.049} & 5274/0.545 \\ \hline
PIE$_{full}$ & \multicolumn{1}{c|}{376/0.045}   & 1167/0.081  & \multicolumn{1}{c|}{402/0.037} & 1636/0.150 & \multicolumn{1}{c|}{338/0.031} & 3547/0.366 \\ \hline
BiTraP       & \multicolumn{1}{c|}{499/0.059}   & 1142/0.079  & \multicolumn{1}{c|}{430/0.039} & 1899/0.174 & \multicolumn{1}{c|}{383/0.035} & 3492/0.361 \\ \hline
BiPed        & \multicolumn{1}{c|}{248/0.029}   & 1049/0.073  & \multicolumn{1}{c|}{277/0.025} & 1077/0.099 & \multicolumn{1}{c|}{269/0.024} & 1526/0.158 \\ \hline
PedFormer    & \multicolumn{1}{c|}{252/0.030}   & 970/0.067   & \multicolumn{1}{c|}{259/0.024} & 924/0.085  & \multicolumn{1}{c|}{248/0.023} & 1422/0.147 \\ \hline
\textbf{ENCORE-D} &
  \multicolumn{1}{c|}{\textbf{222/0.026}} &
  \textbf{877/0.061} &
  \multicolumn{1}{c|}{\textbf{225/0.021}} &
  \textbf{707/0.065} &
  \multicolumn{1}{c|}{\textbf{217/0.020}} &
  \textbf{1055/0.109} \\
\end{tabular}%
}
\end{table}

\begin{table}[h]
\caption{Comparison to SOTA on challenging scenarios using $C_{MSE}$/$sC_{MSE}$ metrics.}
\label{tbl:challenge_cmse}
\resizebox{\columnwidth}{!}{%
\begin{tabular}{l|ll|ll|ll}
             & \multicolumn{2}{c|}{\textbf{Pedestrian State}} & \multicolumn{2}{c|}{\textbf{Ego-motion}}    & \multicolumn{2}{c}{\textbf{Ego-action}}    \\ \hline
 &
  \multicolumn{1}{c|}{\textbf{$W_o$-$S_p$}} &
  \multicolumn{1}{c|}{\textbf{$S_o$-$W_p$}} &
  \multicolumn{1}{c|}{\textbf{Constant}} &
  \multicolumn{1}{c|}{\textbf{Change}} &
  \multicolumn{1}{c|}{\textbf{Straight}} &
  \multicolumn{1}{c}{\textbf{Turn}} \\ \hline
PIE$_{traj}$ & \multicolumn{1}{l|}{513/0.061}   & 1422/0.099  & \multicolumn{1}{l|}{445/0.041} & 2460/0.226 & \multicolumn{1}{l|}{403/0.037} & 4103/0.424 \\ \hline
B-LSTM       & \multicolumn{1}{l|}{487/0.058}   & 1716/0.119  & \multicolumn{1}{l|}{558/0.051} & 3094/0.284 & \multicolumn{1}{l|}{504/0.046} & 5203/0.538 \\ \hline
PIE$_{full}$ & \multicolumn{1}{l|}{359/0.043}   & 1031/0.071  & \multicolumn{1}{l|}{377/0.034} & 1535/0.141 & \multicolumn{1}{l|}{310/0.028} & 3497/0.361 \\ \hline
BiTraP       & \multicolumn{1}{l|}{476/0.057}   & 1028/0.071  & \multicolumn{1}{l|}{404/0.037} & 1795/0.165 & \multicolumn{1}{l|}{354/0.032} & 3425/0.354 \\ \hline
BiPed        & \multicolumn{1}{l|}{231/0.027}   & 905/0.063   & \multicolumn{1}{l|}{252/0.023} & 969/0.089  & \multicolumn{1}{l|}{241/0.022} & 1467/0.152 \\ \hline
PedFormer    & \multicolumn{1}{l|}{234/0.028}   & 836/0.058   & \multicolumn{1}{l|}{233/0.021} & 819/0.075  & \multicolumn{1}{l|}{218/0.020} & 1363/0.141 \\ \hline
\textbf{ENCORE-D} &
  \multicolumn{1}{l|}{\textbf{206/0.024}} &
  \textbf{747/0.052} &
  \multicolumn{1}{l|}{\textbf{200/0.018}} &
  \textbf{603/0.055} &
  \multicolumn{1}{l|}{\textbf{190/0.017}} &
  \textbf{987/0.102} \\ 
\end{tabular}%
}
\end{table}

\begin{table}[h]
\caption{Comparison to SOTA on challenging scenarios using $CF_{MSE}$/$sCF_{MSE}$ metrics.}
\label{tbl:challenge_cfmse}
\resizebox{\columnwidth}{!}{%
\begin{tabular}{l|ll|ll|ll}
             & \multicolumn{2}{c|}{\textbf{Pedestrian State}} & \multicolumn{2}{c|}{\textbf{Ego-motion}}      & \multicolumn{2}{c}{\textbf{Ego-action}}      \\ \hline
 &
  \multicolumn{1}{c|}{\textbf{$W_o$-$S_p$}} &
  \multicolumn{1}{c|}{\textbf{$S_o$-$W_p$}} &
  \multicolumn{1}{c|}{\textbf{Constant}} &
  \multicolumn{1}{c|}{\textbf{Change}} &
  \multicolumn{1}{c|}{\textbf{Straight}} &
  \multicolumn{1}{c}{\textbf{Turn}} \\ \hline
PIE$_{traj}$ & \multicolumn{1}{l|}{2180/0.221}  & 6032/0.360  & \multicolumn{1}{l|}{1916/0.158} & 10677/0.724 & \multicolumn{1}{l|}{1695/0.138} & 18781/1.411 \\ \hline
B-LSTM       & \multicolumn{1}{l|}{2318/0.235}  & 7021/0.419  & \multicolumn{1}{l|}{2466/0.203} & 12909/0.876 & \multicolumn{1}{l|}{2168/0.177} & 23408/1.759 \\ \hline
PIE$_{full}$ & \multicolumn{1}{l|}{1434/0.146}  & 4458/0.266  & \multicolumn{1}{l|}{1520/0.125} & 6473/0.439  & \multicolumn{1}{l|}{1174/0.096} & 16269/1.222 \\ \hline
BiTraP       & \multicolumn{1}{l|}{1973/0.200}  & 4289/0.256  & \multicolumn{1}{l|}{1644/0.135} & 7583/0.515  & \multicolumn{1}{l|}{1399/0.114} & 15340/1.153 \\ \hline
BiPed        & \multicolumn{1}{l|}{862/0.088}   & 3614/0.216  & \multicolumn{1}{l|}{949/0.078}  & 3810/0.258  & \multicolumn{1}{l|}{889/0.073}  & 6174/0.464  \\ \hline
PedFormer    & \multicolumn{1}{l|}{892/0.091}   & 3157/0.188  & \multicolumn{1}{l|}{828/0.068}  & 2956/0.201  & \multicolumn{1}{l|}{776/0.063}  & 4892/0.368  \\ \hline
\textbf{ENCORE-D} &
  \multicolumn{1}{l|}{\textbf{767/0.078}} &
  \textbf{3094/0.185} &
  \multicolumn{1}{l|}{\textbf{717/0.059}} &
  \textbf{2359/0.160} &
  \multicolumn{1}{l|}{\textbf{661/0.054}} &
  \textbf{4222/0.317} \\
\end{tabular}%
\vspace{-0.5cm}
}
\end{table}

The results of comparison to the past arts on challenging scenarios are reported in Tables \ref{tbl:challenge_bmse}, \ref{tbl:challenge_cmse}, and \ref{tbl:challenge_cfmse}. As shown in the tables, across all metrics, the proposed model significantly improves the performance. However, the improvement ratio varies. For instance, the improvement on pedestrian state scenarios is lower in general, since predicting such state changes across observation and prediction horizon requires additional contextual information beyond dynamics, e.g. signal state, pedestrian pose, etc. The improvement in the case of ego-motion scenarios is more prominent, suggesting that the proposed model is modeling ego-motion more effectively, hence, it is better at handling challenging scenarios.

\subsection{Ablation Studies on ENCORE}

\subsubsection{Effect of weighting}

\begin{figure}[!t]
\vspace{0.2cm}
\centering
\includegraphics[width=0.6\columnwidth]{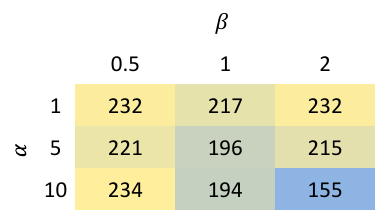}
\caption{The effect of different loss weighting on the overall performance of ENCORE.}
\label{fig:abl_weights}
\vspace{-0.5cm}
\end{figure}
One of the factors in the proposed model, ENCORE, is the mixing weights of loss objectives. In general, larger values are selected for $\alpha$ as the scale $L_{sFT}$ is much smaller compared to other losses. The changes in the performance based on different loss weights are shown in Figure \ref{fig:abl_weights}.

\subsubsection{Sampling}

\begin{table}[]
\caption{The performance of ENCORE using different number of samples, K.}
\label{tbl:abl_k}
\resizebox{\columnwidth}{!}{%
\begin{tabular}{c|c|ccc|c|c|c}
 &    & \multicolumn{3}{c|}{\textbf{$B_{MSE}$}}                       & \textbf{$C_{MSE}$} & \textbf{$CF_{MSE}$} & \textbf{$sB_{MSE}$} \\ \hline
 &    & \multicolumn{1}{c}{\textit{0.5}} & \multicolumn{1}{c}{\textit{1}} & \textit{1.5} & \textit{1.5} & \textit{1.5} & \textit{1.5} \\ \hline
\multirow{6}{*}{\textit{K}} & \textit{3 }& \multicolumn{1}{c}{26}           & \multicolumn{1}{c}{75}         & 181          & 153          & 534          & 0.017 \\ \cline{2-8} 
 & \textit{5}  & \multicolumn{1}{c}{22} & \multicolumn{1}{c}{57} & 131 & 105            & 358             & 0.012  \\ \cline{2-8} 
 & \textit{6}  & \multicolumn{1}{c}{21} & \multicolumn{1}{c}{52} & 119 & 94             & 318             & 0.011  \\ \cline{2-8} 
 & \textit{10} & \multicolumn{1}{c}{18} & \multicolumn{1}{c}{42} & 90  & 67             & 222             & 0.008  \\ \cline{2-8} 
 & \textit{15} & \multicolumn{1}{c}{16} & \multicolumn{1}{c}{36} & 76  & 54             & 177             & 0.007  \\ \cline{2-8} 
 & \textit{20} & \multicolumn{1}{c}{\textbf{15}} & \multicolumn{1}{c}{\textbf{33}}   & \textbf{70}  & \textbf{49}             & \textbf{155}             &\textbf{0.006}  \\
\end{tabular}%
\vspace{-1cm}
}
\end{table}

In the context of multimodal prediction for intelligent driving, depending on the application, different number of samples are drawn, e.g. 5, 10 \cite{yuan2021agentformer}, 6 \cite{nayakanti2023wayformer}, and 20 \cite{halawa2022action}. Different number of samples and the performance measure can reflect the model's confidence, coverage of the amples, etc. We report the results for ENCORE with different number of samples in Table \ref{tbl:abl_k}.

\bibliography{refs}

\end{document}